\documentclass[11pt]{article}

\usepackage[margin=1in]{geometry}
\usepackage{hyperref}
\usepackage{graphicx}
\usepackage{url}
\usepackage{titlesec}
\usepackage{textcomp}
\usepackage{enumitem}
\usepackage{amsmath, amssymb}
\usepackage{booktabs}
\usepackage{array}
\usepackage{caption}
\usepackage{authblk}
\usepackage{longtable}
\usepackage{calc} 
\usepackage{etoolbox}
\usepackage[numbers,super,compress]{natbib}
\usepackage{xcolor}

\title{Atoms to Processes: The Role of Artificial Intelligence and Machine Learning in Chemical Engineering}
\author{Michael Baldea$^a$\thanks{Corresponding author: mbaldea@che.utexas.edu}, Linda J. Broadbelt$^b$, Marianthi G. Ierapetritou$^c$ ,  Akhilesh Jain$^d$, Ankur Kumar$^e$, Thomas A. Kwan$^{f,p}$, F\`{e}lix Llovell$^g$, Andrew J. Medford$^h$, Ilias Mitrai$^a$, Joel Paulson$^n$, Junyi Qiao$^o$, Matthew P. Rivera$^k$, Kirti C. Sahu$^i$,Lev Sarkisov$^m$,
 Zachary P. Smith$^k$, Calvin Tsay$^j$, Ching-Mei Wen$^c$, Victor M. Zavala$^n$,  Huacheng Zhang$^l$, Dan Zhao$^o$\\

$^a$ McKetta Department of Chemical Engineering and Oden Institute for Chemical Engineering and Sciences, The University of Texas at Austin, Austin, TX 78731, USA\\
$^b$ Department of Chemical and Biological Engineering, Northwestern University, Evanston, IL 60208, USA\\
$^c$ Department of Chemical and Biomolecular Engineering, University of Delaware, Newark, DE 19716, USA\\
$^d$ Advanced Analytics and AI, Baker Hughes\\
$^e$ Smart Operations, Global AI, Linde plc, Tonawanda, NY, 14150, USA \\
$^f$ Schneider Electric Research Institute, Schneider Electric, Boston, MA 02108, USA\\
$^g$ Department of Chemical Engineering, ETSEQ, Universitat Rovira i Virgili, 43007 Tarragona, Spain\\
$^h$ School of Chemical and Biomolecular Engineering, Georgia Institute of Technology, Atlanta, GA 30332, USA\\
$^i$ Department of Chemical Engineering, Indian Institute of Technology Hyderabad, Kandi - 502 204, Sangareddy, Telangana, India\\
$^j$ Department of Computing, Imperial College London, London SW7 2AZ, United Kingdom\\
$^k$ Department of Chemical Engineering, Massachusetts Institute of Technology, Cambridge, MA 02139, USA\\
$^l$ Department of Chemical and Biological Engineering, Monash University, Clayton, Victoria 3800, Australia\\
$^m$ Department of Chemical Engineering, The University of Manchester, Manchester M13 9PL, United Kingdom\\
$^n$ Department of Chemical and Biological Engineering, University of Wisconsin, Madison, WI 53706, USA\\
$^o$ Department of Chemical and Biomolecular Engineering, National University of Singapore, Singapore 117585\\
$^p$ Institute for Global Sustainability, Boston University, Boston, MA 02215, USA

 }

\date{\today}

\begin{document}
\maketitle

\newcommand{\cod}{\textnormal{CO\textsubscript{2}}}

\begin{abstract}
The rapid maturation of artificial intelligence (AI) and machine learning (ML) has catalyzed a profound shift in how chemical engineering problems are formulated, analyzed, and solved. Advances in computing, data availability, and learning algorithms have enabled AI/ML methods to impact applications spanning atomic-scale simulations, materials and catalyst discovery, transport and thermodynamics, separations, process systems engineering, and industrial operations. This article provides a  perspective on recent methodological developments and representative applications, emphasizing how AI/ML tools are being integrated with first-principles models to address challenges of predictive accuracy, data scarcity, extrapolation, interpretability, and model lifecycle management.

Across domains, a unifying trend is the move away from purely black-box approaches toward hybrid and physics-informed frameworks that explicitly respect conservation laws, thermodynamic consistency, and known structural constraints. These approaches not only improve robustness and reliability, but also enable meaningful human–AI collaboration by providing information at an appropriate level of abstraction for the task and decision context. We conclude that AI and ML are not replacing the core principles of chemical engineering; rather, they are amplifying them. As the field advances toward increasingly autonomous, adaptive, and sustainable systems, the thoughtful integration of AI/ML with first-principles understanding and domain expertise will be essential to realizing their full potential across both research and industrial practice.
\end{abstract}


\clearpage
\section*{Nomenclature}

\begin{tabular}{@{}p{0.16\linewidth}p{0.78\linewidth}@{}}
AI & Artificial intelligence \\
ANN & Artificial neural network \\
APC & Advanced process control \\
ARIMA & Autoregressive integrated moving average \\
BO & Bayesian optimization \\
CFD & Computational fluid dynamics \\
CNN & Convolutional neural network \\
COF & Covalent organic framework \\
CVAE & Conditional variational autoencoder \\
DAE & Differential-algebraic equation \\
DDPG & Deep deterministic policy gradient \\
DFT & Density functional theory \\
DoE & Design of experiments \\
EoS & Equation of state \\
FAIR & Findable, accessible, interoperable, reusable (data principles) \\
FCC & Fluidized catalytic cracking \\
GNN & Graph neural network \\
GP & Genetic programming; also used for Gaussian process (meaning clear from context) \\
HER & Hydrogen evolution reaction \\
kMC & Kinetic Monte Carlo \\
LHHW & Langmuir--Hinshelwood--Hougen--Watson (kinetic model) \\
LLM & Large language model \\
LSTM & Long short-term memory (recurrent neural network) \\
MAE & Mean absolute error \\
MD & Molecular dynamics \\
MILP & Mixed-integer linear programming \\
MINLP & Mixed-integer nonlinear programming \\
ML & Machine learning \\
MLIP & Machine-learned interatomic potential \\
MMM & Mixed-matrix membrane \\
MOF & Metal--organic framework \\
MPC & Model predictive control \\
ODE & Ordinary differential equation \\
ORR & Oxygen reduction reaction \\
PCA & Principal component analysis \\
PdM & Predictive maintenance \\
PINN & Physics-informed neural network \\
PLS & Partial least squares \\
PSE & Process systems engineering \\
QSPR & Quantitative structure--property relationship \\
RL & Reinforcement learning \\
RNN & Recurrent neural network \\
RTO & Real-time optimization \\
SDL & Self-driving laboratory \\
SINDy & Sparse identification of nonlinear dynamics \\
SVM & Support vector machine \\
UQ & Uncertainty quantification \\
VAE & Variational autoencoder \\
VSA & Vacuum swing adsorption \\
\end{tabular}

\clearpage
\section{Introduction}
The adoption of machine learning (ML) and artificial intelligence (AI) tools and methods in chemical engineering research and applications has been under discussion since at least the 1990s. However, widespread translation of these methods into industrial practice did not occur in the subsequent decades. Venkatasubramanian \cite{venkatasubramanian2019promise} identified several contributing factors: the intrinsic difficulty of many problems, the availability of satisfactory solutions through conventional approaches, and the limited computational capabilities of the time. Further difficulties came from the fact that many ML/AI methods inherently require large amounts of data for model training and validation.

Since then, the substantial advance in computing power (boosted by specialized hardware such as GPUs), data acquisition and storage, coupled with the capacity to train large-scale ML models, have enabled transformative progress across diverse domains, including e-commerce, computer vision, natural language processing, and, more recently, generative AI.

In the scientific and engineering context, high throughput experimentation as well as new modeling and computation tools accelerated the generation of data, and ML/AI tools have already demonstrated utility in applications such as catalyst discovery and prediction of physicochemical properties of organic molecules.

The effective deployment of ML and AI models in science and engineering requires careful attention to several specific considerations. First, predictive accuracy is paramount. For example, errors in predicting the viscosity of a novel molecule at the computational design stage could produce a solid at the synthesis stage, whereas the target was a liquid material. At the process level, overestimation of product purity may result in the production of off-specification material and associated penalties, while underestimation may yield unnecessarily high-purity product and associated revenue loss.
In this context, it can be equally important to quantify the uncertainty associated with model predictions, as awareness of prediction confidence can guide decision-making, risk mitigation, and additional validation. In the above example, a model predicting a liquid with high uncertainty might trigger further validation before synthesis.

Second, ML models must remain consistent with physical laws, as the data on which they are trained represent physical systems constrained by mass and energy conservation laws. Third, models should exhibit a degree of interpretability, allowing their behavior to be understood in relation to the governing equations and principles of the underlying system. Fourth, interpretability should allow for transfer learning, or the ability of a model to represent systems that are physically/structurally similar to those whose data it was trained on, potentially with some amount of fine tuning. This is essentially an extrapolation capacity. Fifth, model maintenance is essential. Thinking particularly at the macroscopic scale, industrial plants evolve continuously: equipment replacement, catalyst deactivation, corrosion, and fouling all lead to gradual deviations between model predictions and plant performance. Accordingly, models must be updated regularly to mitigate prediction drift.

Last but not least, reproducibility is a precondition for all of the above, yet it remains one of the field's most persistent and under-addressed weaknesses. Reported model performance can be verified, compared fairly across studies, or built upon only when the underlying computational workflow is fully disclosed and re-runnable. In practice, chemical engineering AI/ML publications routinely omit one or more of the following: source code, data preprocessing and featurization scripts, the exact train/validation/test splits, trained model weights, hyperparameter settings and the search procedure used to obtain them, random seeds, and the software and library versions  needed to regenerate results. The absence of these materials has repeatedly forced researchers to spend substantial effort reimplementing or guessing at published workflows, and it makes it difficult to distinguish genuine methodological advances from artifacts of undisclosed data splits, tuning budgets, or preprocessing choices. Adjacent computational disciplines and journals, e.g., the Journal of Chemical Information and Modeling, have moved toward explicit minimum reproducibility requirements for computational submissions. We argue that chemical engineering AI/ML research, and the journals and venues that publish it, should move toward comparable, explicitly stated minimum standards for code and data availability and computational reproducibility.

This contribution outlines the open challenges associated with these requirements and discusses promising avenues for advancing the application of AI and ML in chemical engineering. Following a high-level overview of recent relevant methodological developments, we organize the paper along the sections of \textit{Industrial \& Engineering Chemistry Research} (I\&ECR).

\section{Engineering-Relevant AI/ML Methodology}

The developments and challenges reviewed above emphasize the breadth and diversity of problems encountered across chemical engineering, which has naturally led to the adoption of a broad range of AI and ML methods. These methods differ in how they utilize data, incorporate domain/physical knowledge, and balance interpretability with predictive power.

The development of AI and ML models often begins with careful formulation of the learning task: regression or classification for property prediction, sequence modeling for time-series plant data, or generative modeling for molecular and process design. A critical first step involves representing chemical and engineering data and domain knowledge as mathematical inputs. Examples include structured fingerprints or graph representations for molecules, topology- or string-based representations for reactions, and digitized process flowsheets for plant models \cite{dobbelaere2021machine,haghighatlari2020learning}. Data reconciliation is a common step in  engineering because raw data may violate mass/energy balances, contain noise, or reflect operating constraints that must be preserved.

\begin{figure}[htb]
\includegraphics[width = \textwidth]{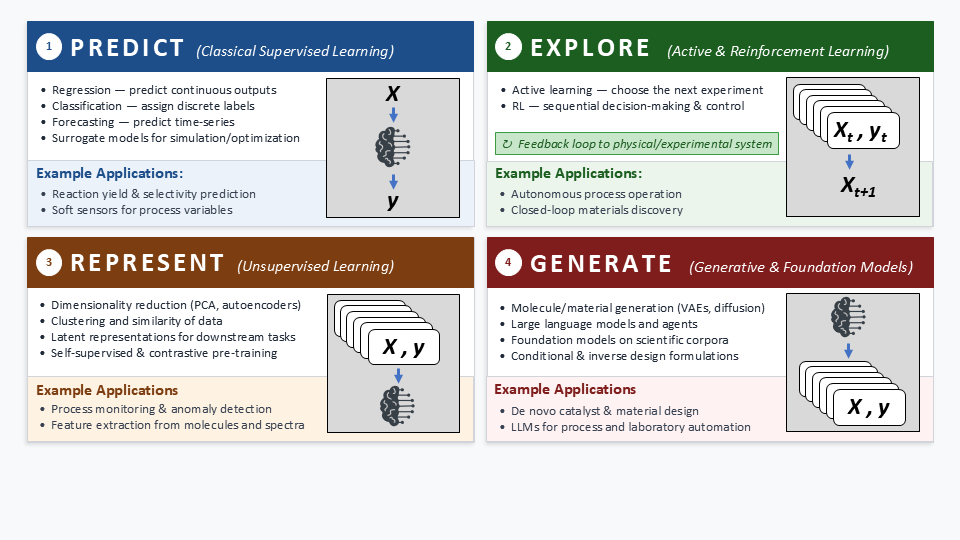}
\caption{AI/ML methodological classes relevant to chemical engineering applications}
\label{fig:classes}
\end{figure}

Below, we highlight several methodological classes that have proven particularly relevant to chemical engineering applications (Figure \ref{fig:classes}).

\subsection{Classical Supervised Learning: Regression, Forecasting, and Classification}

Supervised learning remains the dominant paradigm for chemical engineering applications of AI. In this setting, models are trained on labeled data pairs to predict either continuous outputs (regression) or discrete labels (classification). This setting occurs in property prediction, surrogate modeling of high-fidelity simulations, and general modeling of experimental data. Classification algorithms are commonly employed for fault detection and diagnosis, quality monitoring, and operating-regime identification \cite{taqvi2021review}. Regression models range from simple, e.g., linear and kernel-based approaches, to deep neural networks, reflecting a trade-off between expressivity, interpretability, and data requirements \cite{haghighatlari2020learning}. Linear and generalized linear models remain widely used because they offer mechanistic interpretability and naturally align with many engineering correlations, especially when the input features are designed strategically \cite{mowbray2022industrial}.

Kernel-based methods and support vector machines provide nonlinear modeling capacity while remaining simple in form, making them effective for the moderate data volumes found in experimental areas. Ensemble methods, such as random forests and gradient boosting, can offer improved accuracy without significant additional model design. The advent of neural networks and deep learning architectures, including fully connected networks, convolutional networks, and graph neural networks, saw a significant increase in chemical engineering applications, spanning molecular, material, and process scales. Methodologically, while deep learning models can be universal approximators, a significant challenge in chemical and engineering applications is controlling overfitting with limited data \cite{thebelt2022maximizing}; several paradigms are used to mitigate this risk, including regularization terms \cite{tsay2021sobolev}, judicious architecture design \cite{rittig2023graph}, data augmentation \cite{lee2021datadriven}, and hybrid modeling \cite{bradley2022perspectives}.

Of the above, hybrid modeling has seen a particular focus in the literature \cite{psichogios1992hybrid}. Chemical engineering data obey physical laws, e.g., mass and energy conservation, thermodynamic constraints, kinetic rate laws, and known symmetries. Hybrid models seek to integrate data-driven models within these first-principles equations, e.g., using ML to estimate unknown relationships, refine thermodynamic or kinetic parameters, or provide corrections to the outputs of mechanistic simulators \cite{bradley2022perspectives}. These approaches maintain and incorporate known physical knowledge while allowing the data-driven component to approximate unknown relationships between the input variables, producing models that are more interpretable and better behaved under extrapolation than purely black-box learners. For many chemical engineering tasks, hybrid supervised learning represents a promising compromise between empirical correlations and fully data-driven models. More recently, ``physics-informed'' ML approaches embed physical structure or relationships directly into architectures or loss functions. Methodologies here include designing model connectivity based on known relationships \cite{alhajeri2022physicsinformed}, including physics-based penalties, such as governing equations and/or optimality conditions, during training \cite{choi2022physicsinformed, agyeman2026hybrid, nellikkath2022physics, chen2024physics}, and physics-based feature engineering \cite{chew2024advancing}. In general, these inductive biases help restrict the space of model parameters to be physically feasible, improving model performance, as well as interpretability and extrapolation safety.

A subfield of supervised learning focuses on forecasting, or making predictions based on historical time-series data. Building these ``dynamical system'' models (i.e., system identification), is a classical area of study in the field of process control, leading to methods such as state-space models, Hammerstein--Wiener models, and autoregressive integrated moving average (ARIMA) models. These classical models are relatively compact and amenable to deployment in linear model predictive control \cite{muske1993model}, nonlinear control \cite{hernandez1993control}, and optimization-based scheduling \cite{caspari2020integration}. More recent work has focused on deep-learning forecasting models, such as recurrent neural networks and transformer models, which are capable of learning more complex dynamics, such as unknown reaction pathways \cite{ji2021autonomous} and crystallization kinetics \cite{sitapure2023exploring}. Finally, an orthogonal line of research has sought dynamic symbolic models, providing interpretability and potentially helping uncover mechanistic equations directly from data \cite{lejarza2022datadriven,lejarza2023dynamic}.

\subsection{Active Learning and Reinforcement Learning}

Data are often expensive to generate, whether from process operations, bench-scale experiments, or computational simulations \cite{thebelt2022maximizing}. Active learning seeks to maximize information gain from limited data acquisition by iteratively selecting new samples where the model uncertainty is highest. Traditional approaches relied on model-based design of experiments \cite{franceschini2008modelbased}, system identification, or dual control. More recently, ML- and AI-based design-of-experiments (DoE) have demonstrated promise in materials discovery, reaction optimization, and process calibration, where automated and autonomous experimental platforms can reduce both cost and time \cite{schweidtmann2018machine}. A particularly influential class of algorithms here is Bayesian optimization (BO), which views experiment design from the lens of global optimization of an expensive black-box function \cite{paulson2025whyyour,paulson2025bayesian}. BO couples a probabilistic ML model with an acquisition function that quantifies the value of experiments, i.e., to balance exploration and exploitation \cite{frazier2018bayesian}. To tackle chemical engineering challenges, methodological advances have sought to integrate BO in high-dimensional or hybrid (discrete/continuous) variable settings \cite{thebelt2022maximizing}, BO with multi-fidelity data (e.g., combining experiments and simulations) \cite{savage2024machine}, and BO with hybrid, grey-box models \cite{paulson2022cobalt}, enabling more reliable optimization when training data are sparse or partially simulated.

Whereas active learning focuses on data acquisition, a separate paradigm, reinforcement learning (RL) addresses the problem of sequential decision-making. In RL, an agent learns optimal control strategies directly through interaction with an environment. This formulation naturally aligns with process systems engineering challenges, including control, scheduling, adaptive experimentation, and process optimization tasks \cite{bloor2025survey,devarakonda2025recent}. In particular, modern deep RL algorithms (e.g., deep deterministic policy gradients, proximal policy optimization) provide increasing levels of model complexity and flexibility for nonlinear, high-dimensional chemical engineering systems \cite{spielberg2019toward}. Nevertheless, exploration is often hazardous or impractical in these applications, motivating the use of hybrid RL strategies. Two main paradigms have emerged here: (1) learning from simulated data, reduced-order models, or off-policy data from an expert controller \cite{kim2020modelbased,yoo2021reinforcement}, and (2) hybrid frameworks integrating physical knowledge, constraints, or controllers to enforce safe behavior \cite{bloor2024control,burtea2024constrained,lawrence2022deep}.

\subsection{Unsupervised Learning: Dimensionality Reduction and Embeddings}

Unsupervised learning methods uncover underlying structure (i.e., a subspace) in high-dimensional data without requiring labeled outputs and are central to multivariate process monitoring in chemical engineering. Classical latent-variable methods such as principal component analysis (PCA) and its (dynamic) extensions can project correlated process or spectroscopy data for later ML applications, e.g., soft sensors \cite{dong2018novel,qin2006overview,srinivasan2004dynamic}. More recent works apply ML models such as autoencoders, variational autoencoders (VAEs), and geometric manifold learning techniques to learn more complex latent spaces underlying data. At the molecular and materials scale, representation learning plays an analogous role. Variational autoencoders and related models learn continuous molecular embeddings to define chemically similar molecules \cite{cheng2020mapping}; this has enabled inverse materials design workflows by optimizing in continuous latent space and decoding to valid structures \cite{gomezbombarelli2018automatic,griffiths2020constrained}.

\subsection{Generative Modeling and Foundation Models}

Generative models learn mappings from low-dimensional latent spaces to distributions over molecules, materials, or process systems, thereby shifting the focus from prediction to design. This is essentially the inverse task of prediction models: sampling candidates that satisfy desired properties rather than predicting properties. For example, the above studies \cite{gomezbombarelli2018automatic,griffiths2020constrained} use VAEs for molecular design, while Yue et al.\ \cite{yue2025benchmarking} provide a benchmarking comparison of generative models for polymer design. Following the advances in the ML and AI fields, more complex models such as GPT- and diffusion-based models are recently employed for similar generative tasks, e.g., in polymer design \cite{yang2024denovo} and process flowsheeting \cite{vogel2023learning}.

In parallel, the emergence of foundation models, i.e., large, pre-trained architectures adaptable to downstream tasks, has begun to reshape workflows across chemical engineering fields. Within foundation models, a particular focus has been large language models (LLMs), or pre-trained models on natural language that can directly be used, e.g., for documentation and education tasks \cite{tsai2023exploring}. Many recent works adapt LLMs to the ``language'' of chemical engineering, potentially accelerating the pace of discovery \cite{dagdelen2024structured,ramos2025review}. Specialized models such as Molformer \cite{wu2023molformer} and Chemformer \cite{irwin2022chemformer}  provide foundation models pre-trained for chemistry-related tasks, while tabular foundation models \cite{hollmann2023tabpfn,hollmann2025accurate} leverage in-context learning to make predictions using small datasets (characteristic of many chemical engineering problems) without task-specific training.

Beyond passive generation, recent work also explores tool-augmented and agentic AI systems, in which foundation models are coupled with external tools such as process simulators, optimization routines, and scientific databases to iteratively solve complex tasks. In this paradigm, large language models (LLMs) can orchestrate complex, multi-step workflows by generating inputs to simulation environments, invoking property prediction models, or querying experimental data~\cite{lima2026improving,rupprecht2026multi}. Predictions can be improved based on returned results, inter-agent coordination, and/or interaction with the human user. Early demonstrations include automated drug discovery and materials design~\cite{m2024augmenting,taskiran2025knowledge}, control and operational decision-making~\cite{pajak2025multi,vyas2025autonomous}, and process design and simulation~\cite{bahamdan2026sketch2simulation,schafer2026context}. While still emerging, this direction is particularly relevant to chemical engineering, where established first-principles models and simulation environments provide a natural interface for hybrid, AI-driven decision-making systems that bridge data-driven and physics-based approaches.


\section{Kinetics, Catalysis and Reaction Engineering}

Kinetics, catalysis, and reaction engineering have been significantly influenced by ML and AI models in recent years. The use of ML in catalysis dates back to the early days of ML and AI in chemical engineering, and these early impacts are reviewed elsewhere\cite{medford2018extracting}. More recently, ML models have become ubiquitous in the field of computational catalysis, where atomic-scale and microkinetic models are used to understand and predict catalytic materials. In addition, ML models have been used extensively in conjunction with or in place of traditional optimization techniques to extract kinetic equations and rate parameters from data measured using real or simulated chemical reactors. Finally, ML and AI models have been heavily used to discover and optimize catalyst formulations, replacing or augmenting traditional ``quantitative structure-property relationships'' and accelerating high-throughput computational and experimental screening. Here, we briefly review some recent representative examples of each area, focusing on primary sources.

\subsection{Simulations of catalysts and kinetic processes}
\label{sec:simulations-of-catalysts}
\subsubsection{Atomic-scale simulations of active sites and surfaces}
\label{sec:atomic-scale-simulations}

The use of atomic-scale models to understand catalytic processes is a common strategy in the field, and has historically been achieved using density functional theory (DFT), which captures covalent bonding effects critical to catalysis, but is fast enough to be applied to surface slabs commonly used to model solid/liquid or solid/gas interfaces (see, e.g., \cite{norskov2011density}). Even though DFT is faster than other quantum-mechanical methods, it is still far too slow for realistic catalyst particles or for the large-scale sampling needed for statistical-mechanical analyses. The advent of machine-learned interatomic potentials (MLIPs) has been transformative to the field of computational catalysis, as they provide accuracies comparable to DFT ($\sim 0.2$--$0.5$ eV errors on adsorption energies) at speeds thousands of times faster than DFT (e.g., \cite{chanussot2021open}). Combining these MLIPs with DFT through active learning has become ubiquitous in the field of computational catalysis, and is now widely used to accelerate common tasks like geometry optimizations and transition-state searches (e.g., \cite{garijodelrio2020machine,peterson2016acceleration}). One notable example is AdsorbDiff, which uses a generative diffusion model to directly predict adsorbate configurations \cite{kolluru2024adsorb}. In addition, MLIPs have been used in conjunction with DFT to find global minima of various types of adsorbates and active sites \cite{peterson2014global,kaappa2021global,jung2023machine}, a task that is computationally intractable with DFT alone. Moreover, MLIPs have enabled substantial advances in the statistical sampling of potential energy surfaces of adsorbed species, providing a route to compute more accurate entropies and free energies of adsorption and activation \cite{stocker2023estimating,gupta2024kinetic}. Finally, MLIPs have drastically increased the length scales that are accessible in catalytic simulations, making it possible to model entire supported nanoparticles \cite{maxson2025metal} or micron-scale surfaces \cite{johansson2022micron}. These AI and ML capabilities have substantially improved the complexity of active site models and the accuracy of surface reaction free energies that are critical to bottom-up simulations of catalyst surfaces. However, it is worth considering some existing limitations of MLIPs. These approaches are only ever capable of reproducing the accuracy of the underlying electronic structure methods used to generate training data, and even these methods are known to suffer from accuracy limitations \cite{Mardirossian_2017}. Moreover, validation of long timescale simulations with MLIPs is difficult, since it is not possible to re-run the full simulation at DFT-level fidelity. Spot checks along the trajectory are often used to assess validity, and active learning schemes can be used to combine MLIPs with DFT to improve accuracy \cite{Vandermause_2020,garijodelrio2020machine,peterson2016acceleration}. Uncertainty quantification for MLIPs is an emerging area of interest, although assessing the quality of uncertainty estimates is challenging, especially outside the training domain \cite{Tran_2020,Hu_2022,Zhan_2021,Pernot_2022}. Nonetheless, MLIPs are one of the most impactful AI/ML developments in computational materials science and catalysis \cite{chanussot2021open,tran2023open,Medford_2026}.

\subsubsection{Identification of reaction networks and mechanisms}

Beyond active sites, it is also critical to consider reaction networks when modeling catalytic reactions. Chemistries of interest in kinetics and catalysis typically consist of thousands of reactions and species, particularly when reactive intermediates are included. If simulations of the complex chemistry of thermal and catalytic processes are to be carried out with high fidelity, it is critical that tools for the generation of mechanisms, active sites and microkinetic models be developed and applied, which is the purview of AI and ML. Early efforts trace back to the 1970s, in which computers were armed with information about patterns of reactivity, now referred to template-based approaches, gleaned from experts and organic chemistry textbooks, an early form of AI. Various groups have developed software for mechanism generation over the years (e.g., \cite{broadbelt1994computer,vandewiele2012genesys}), and a network or mechanism generator has become an essential part of a kinetic modeler’s toolbox (e.g., \cite{horton2015molecular}). The earliest AI tools for mechanism generation focused on homogeneous chemistries such as pyrolysis and combustion, but more recent efforts have tackled the complexity engendered by a catalytic surface, active sites and adsorbates. In the realm of generating complex reaction networks for reactions on heterogeneous catalysts, recent examples include RMG-Cat \cite{goldsmith2017automatic} and Genesys-Cat \cite{ureel2025genesyscat}. RMG-Cat iteratively proposes new reactions according to reaction templates, but it includes only the reactions that are kinetically significant based on a precompiled database of thermodynamic properties and rate coefficients for known species and reactions, along with the ability to estimate these parameters for new species and reactions. To validate it, the authors showed that it successfully ``discovered'' many of the same reactions as a previously compiled mechanism that was developed over many years by experts for an example reaction system. Genesys-Cat operates in a similar fashion, functioning as both a model generator that can generate gas- and surface-phase mechanisms in parallel, which are compatible with standard packages such as Chemkin and Cantera, demonstrated well on catalytic cracking of iso-octane for three different zeolites but also being applicable to conventional metal catalysts. A challenge that arises for reactions on heterogeneous catalytic surfaces, however, is determining where adsorbates bind and how neighboring adsorbed molecules affect their adsorption kinetics and thermodynamics. Kolluru and Kitchen address this challenge with machine learning \cite{kolluru2024adsorb}, which not only provides guidance for development of microkinetic models for a given catalytic surface or facet, but also allows for high throughput evaluation. As the authors note, the quest for the lowest energy adsorbed geometries has relied on heuristics, problem-specific intuitions, or brute-force approaches to guide adsorbate placement. Instead, they propose a novel framework for adsorbate placement using denoising diffusion which includes optimization with a pretrained machine learning force field, the output of which can be analyzed using density functional theory.

Estimation of rate coefficients remains the most significant challenge of complex kinetic modeling, as a rate coefficient must be estimated for every elementary step reaction, adsorbate configuration, surface facet, and active site. Here again, though, ML is being deployed to great benefit. As a recent example, Zhang et al.\ \cite{zhang2025enhancing} studied intramolecular H-migration in ether peroxy radicals, species that are relevant to low-temperature combustion. A machine learning framework was developed based on two topological descriptors, capturing reactant topology, and five reaction-specific descriptors, characterizing reaction center properties, to estimate rate coefficients. It was shown that the ML model outperformed rate rule approaches, particularly for cases that were not easily generalizable, with significantly enhanced computational efficiency. Shi et al.\ \cite{shi2022supervised} showed similar success with a ML model used to predict rate coefficients for the interaction of ozone with various organic contaminants. While the ML models were not readily interpreted mechanistically, their practical application and ability to fit well with workflows focusing on complex mechanism generation are notable. This was also the focus of the work of Liu et al.\ \cite{liu2025machine}, who proposed a framework that employs ML to accelerate the analysis of chemical reaction networks for complex gas-phase reaction systems. In addition to using ML to estimate rate coefficients, associated prediction uncertainties of reactions within the chemical systems and influential reactions with high-level uncertainty of predicted rate constants are identified.

\subsection{Treatment of coverage effects in kinetic models}

Another area where AI and ML models are having a growing impact on kinetic modeling is in the treatment of lateral adsorbate–adsorbate interactions. These interactions generate strong coverage dependence in adsorption energies and reaction barriers, often reshaping dominant pathways. They are notoriously difficult to map with DFT because of the combinatorial explosion of possible surface configurations. To address this challenge, Lym et al.\cite{lym2019latticecnn} introduced a lattice convolutional neural network capable of learning configuration- and coverage-dependent adsorption energies directly from DFT data, providing a fast surrogate that can be embedded into lattice kMC or microkinetic models. Building on the idea of correcting mean-field approximations, Tian and Rangarajan\cite{tian2021coveragecorrection} showed that neural networks can learn systematic, coverage-dependent corrections that reconcile mean-field microkinetics with lattice kMC in the fast-diffusion limit. More recently, Johnson et al.\cite{johnson2025coverage} combined automated DFT workflows with machine-learning models to predict coverage-dependent rate constants for multi-step reaction networks, enabling microkinetic simulations that explicitly account for co-adsorbate effects. Another example is provided by Schwalbe-Koda et al.\cite{schwalbekoda2025comprehensive}, who developed a neural-network potential that enabled exhaustive sampling of millions of high-coverage surface configurations for CO and CO/CHOH co-adsorption, revealing how lateral interactions and site blocking reshape adsorption landscapes and reaction energetics. Sakai et al.\cite{sakai2025ethylene} used a similar strategy of sampling high-coverage configurations with a neural network potential to evaluate coverage effects in ethylene hydroformylation, coupling the high-coverage sampling and models with descriptor-based catalyst screening to evaluate the trends in coverage effects across different metals . Interestingly, Klumpers et al.\cite{klumpers2022lateral} used ML models to show that the dynamics of high-coverage adlayers are also important, using neural networks trained on DFT/MD data to extract lateral interaction energies in dynamic, finite-temperature adlayers, leading to strong agreement with experimental measurements. In a complementary strategy, Yokaichiya et al.\cite{yokaichiya2024onthefly} directly embedded neural network models into an adaptive kMC simulation, implicitly capturing lateral interactions during a kMC simulation.

These studies demonstrate that ML approaches can now resolve coverage effects at scales and levels of configurational detail that are impractical with DFT alone, offering a viable path toward more realistic and predictive kinetic models.

\subsection{Integration of AI/ML with multiscale and reactor models}
\label{sec:integration-of-aiml-with-multiscale-and-reactor-models}

The use of AI and ML models in reaction engineering and multiscale modeling is also increasing. Several studies focus on constructing ML surrogates for first-principles microkinetics, enabling reactor-scale simulations that retain mechanistic fidelity while dramatically reducing computational cost. For example, Biermann et al. \cite{biermann2025physicenhanced} developed physics-enhanced neural networks that emulate a UBI-QEP microkinetic mechanism inside industrial packed-bed reactor calculations, while Döppel and Votsmeier \cite{doppel2022efficient}built neural-network–based surrogates for rate-determining steps in first-principles Pt oxidation kinetics. Related work by Bracconi et al.\ and by Partopour et al.\ examines how ML models can rapidly approximate microkinetic rate maps or translate mechanistic reaction networks into efficient surrogate models for reactor design studies \cite{bracconi2020training,partopour2018random}. These approaches enable the integration of detailed kinetics with mass and heat transfer models that would be intractable using standard numerical methods.

 Other efforts aim to use AI and ML methods to automate the entire multiscale workflow from DFT to reactor, such as the AMUSE framework of Sabadell-Rend\'{o}n et al.\ \cite{sabadellrendon2023amuse}, which constructs reaction networks and reactor simulations directly from ab initio energetics, and the work by N\'{u}\~{n}ez and Vlachos\cite{nunez2019activelearning}, which uses active learning to explore how nanoscale catalyst microstructure affects reactor-scale performance without any empirical parameter tuning.

  Together, these works illustrate how ML can provide computational acceleration, automated model construction, and systematic exploration within fully mechanistic multiscale reactor frameworks.

\subsection{Extracting kinetic parameters from measured reactor data}


Once kinetic models based on complex reaction mechanisms have been built, AI and ML play an important role in confronting the results of the model with experimental data through optimization, creating more parsimonious models, or even replacing the bottom-up approach of building a kinetic model based on elementary steps by extracting mechanism or rate laws directly from experimental data. P\'{e}tremand et al.\ \cite{petremand2025methanation} posed the question of whether machine learning (ML)-based regressions can reliably describe kinetics without requiring detailed reaction mechanisms. A data set spanning various temperatures, pressures and space velocities for CO\textsubscript{2} methanation on a Ni-based catalyst was employed to perform several ML-based regressions, the best of which were selected and assessed against two standard kinetic modeling approaches: power law and Langmuir–Hinshelwood–Hougen–Watson (LHHW). The ML-based models outperformed the power law and were comparable to the LHHW model, particularly when tested against data obtained from a nonisothermal reactor. Gusm\~ao et al. \cite{gusmao2023kineticsinformed} put forth the concept of kinetics-informed neural networks, where feed-forward artificial neural networks are used as basis functions to solve systems of DAEs, demonstrating that the simultaneous training of neural nets and kinetic model parameters leads to the solution of the inverse problem through the estimation of kinetic parameters from synthetic experimental data.

\subsection{Using ML to extract kinetic equations}


A related direction is the use of ML models to extract the form of the kinetic equations that underlie kinetic data. Brunton et al.\ \cite{brunton2016sindy} developed an approach termed the Sparse Identification of Nonlinear Dynamics (SINDy) methodology, which sparked many derivatives \cite{lyu2025doesindy} and similar approaches. For example, Prabhu and coworkers \cite{prabhu2025derivativefree} developed an approach to learn interpretable dynamical models from data to handle inherent experimental noise by employing a new derivative-free technique that approximates the integral rather than the derivative and including domain information such as mass balance and chemistry information. Towards this same end, Van Dam and coworkers introduced Poly-ODENet \cite{wu2022polyodenet} to address the challenge of discovering the rate laws (i.e., the ``right-hand side'' of a set of ODEs that describe kinetics of a reaction network following mass action) without knowing the reaction network, especially when the data are incomplete due to experimental limitations. Lejarza et al.\ \cite{lejarza2022discovering,lejarza2023dynamic} take this a step further to address nonlinear systems in an integrated approach that obtains both rate laws and the associated parameters. Given a large set of candidate elementary reaction steps and experimental or simulated concentration data, they find the governing microkinetic model by simultaneously estimating the associated kinetic parameters (e.g., rate constants, activation energies, pre-exponential factors) and systematically eliminating insignificant/redundant elementary reaction steps. A common theme in many of the physics-informed machine learning for kinetic models is the use of known constraints. Although the details vary by model, most ML approaches for fitting kinetics exploit the known structure of the underlying differential equations by constraining the optimization to follow the law of mass action \cite{wu2022polyodenet, prabhu2025derivativefree}. In addition, constraints related to site conservation, thermodynamic consistency, reaction stoichiometry and explicitly known reaction mechanisms have been used to restrict the model space to physically plausible models \cite{Kircher2024, prabhu2025derivativefree, gusmao2023kineticsinformed, lejarza2022datadriven}.

\subsection{Simultaneous model/experiment optimization of reaction conditions}

Models play a critical role in the optimization of reaction conditions to realize optimal yields and selectivities, and ML is playing an increasingly significant role in the design loop for reactive processes. For example, Yang and coworkers \cite{yang2024methanation} developed interpretable machine learning to accelerate the reverse design and optimization of methanation catalysts. To elucidate the intricate interplay among catalyst compositions, preparation conditions, reaction parameters, and catalytic performance, the authors developed different ML models, carried out an interpretable analysis to ascertain the most significant features, and then integrated the best performing model with an optimal multiobjective optimization algorithm to facilitate the reverse design of a low-temperature and highly active catalyst for CO\textsubscript{2} methanation. Neumann and Palkovits \cite{neumann2022reinforcement} examined reinforcement learning approaches for the optimization of the partial oxidation of methane. Q-learning (QL) agents and deep deterministic policy gradient (DDPG) agents were trained to maximize hydrogen production by partial oxidation of methane, with the DDPG agent performing well and demonstrating how reinforcement learning can address a classic reactor optimization problem by adjusting temperature, pressure, flow velocity, and substrate composition. Zhang et al.\ \cite{zhang2022fcc} took on the challenge of applying an ensemble-learning approach to a classic chemical engineering operation, fluidized catalytic cracking (FCC), and showed a learning model can accurately predict the coke yield, noting that the model structure is simple and easy to construct, which provides performance advantages in FCC process modeling.

Machine learning has also shown value in studying processes that are less mature, including optimizing hydrogen production with Fe/g-C\textsubscript{3}N\textsubscript{4} photocatalysts \cite{arabaci2025predictive}, and electrochemical reactors \cite{luo2022electrochemical}. Indeed,  Luo et al.\cite{luo2022electrochemical} note that, because of the complexity of its reaction mechanism and the largely unknown electron transfer pathways, the development of a first-principles-based operational model of a CO\textsubscript{2} electrocatalytic reactor is still in its infancy, and ML thus has an important role to play. Sheng et al.\ \cite{sheng2024autonomous} demonstrate this as well, showing how an autonomous electrochemical platform that implements an adaptive, closed-loop workflow for mechanistic investigation of molecular electrochemistry can discern an electrochemical mechanism amid negative controls and outliers, adaptively design desired experimental conditions, and quantitatively extract kinetic information, from which mechanistic insights were gained, illustrating how ML can enable self-driving electrochemistry laboratories. In many of these cases, experiments themselves serve as validation of the models, either by evolving the model through active learning schemes, or by serving as a final validation of a model prediction.

Tools that enable researchers in the field to more easily adopt ML into their own workflows for kinetic analysis are being developed, as well as data benchmarks for chemical reactions~\cite{boyne2025catechol, felton2021summit, wigh2024orderly}.
For example, Chemfit, developed by Jiscoot and coworkers, facilitates the evaluation of the experimental data quality and quantity requirements needed for the reliable discovery of the rates driving corresponding chemical models \cite{jiscoot2023modelbased}. While conversion of mechanisms in human-readable form into model files has long been a staple of automated network generation and associated kinetic model toolboxes, human/AI collaboration is taken one step further by leveraging LLM prompts to facilitate the process. For example, Li et al.\ introduce a workflow and open-source Python package, the Sanofi Kinetic AI (SKAI) tool \cite{li2025mechanistic}, that simplifies kinetic modeling. SKAI incorporates a prompt-engineered large language model (LLM) that converts reaction descriptions into system equations.

\subsection{Catalyst discovery and optimization}

\subsubsection{Accelerating high-throughput atomic-scale computational screening}
\label{sec:Accelerating-high-throughput-atomic-scale-computational-screening}

Leveraging the modeling developments reviewed in Section \ref{sec:simulations-of-catalysts}, AI and ML are also accelerating the discovery of \textbf{new} catalytic molecules and materials by enabling high-throughput exploration of large compositional and structural spaces that would be prohibitively expensive to evaluate with DFT alone.

In heterogeneous catalysis, the Open Catalyst Project has established large-scale DFT datasets and benchmark tasks that support ML models capable of rapidly predicting adsorption energies for diverse surfaces, enabling orders-of-magnitude faster preliminary screening \cite{chanussot2021open,tran2023open}. Complementary work by Li et al.\ \cite{li2017bimetallic} and Mao et al.\ \cite{mao2021alloynanoclusters} extends these ideas to bimetallic alloys and alloy nanoclusters, using neural-network or descriptor-based surrogates trained on DFT to target improved oxygen reduction reaction (ORR) or hydrogen evolution reaction (HER) activity across vast design spaces. Work by Pisal et al.\ expands on these ideas by using ML models to incorporate the influence of different sites and facets in the exploration of CO\textsubscript{2}-to-methanol catalyst compositions \cite{pisal2025machine}. Similar workflows have emerged for other classes of catalytic materials. For example, Zhang et al.\ employed multitask learning to predict multiple stability metrics across thousands of MOFs \cite{zhang2025mofmultitask}. On the molecular catalysis side, ML models trained on DFT-computed properties have been used to virtually screen redox-active transition-metal complexes, as demonstrated in work by Janet et al.\ \cite{janet2017leveraging}. These efforts illustrate how ML-driven surrogates and learned descriptors can dramatically expand the chemical search space accessible to computation, enabling more efficient identification and optimization of promising catalyst candidates. In many of these cases, the final validation occurs by comparison with DFT simulations, rather than catalyst synthesis and testing. This is appropriate as a test of the ML models because it is not possible for them to exceed the quality of the data they were trained on, and there are many complexities that can cause real synthesized catalysts to have active sites different from those of the underlying atomistic models. These problems are typically tackled with additional types of AI/ML models such as response surface modeling and autonomous synthesis and testing platforms as described below.


Beyond high-throughput atomistic modeling, the approach of response-surface modeling has emerged as a powerful tool for experimental catalyst discovery and optimization, enabling systematic exploration of multidimensional formulation spaces that are difficult to navigate by intuition alone. In recent work, Uusitalo et al.  used a response-surface model along with design-of-experiments to optimize bimetallic compositions for EMF hydrogenation, revealing nonlinear trade-offs between metal ratios and selectivity \cite{uusitalo2022systematic}. In a related work, QSPR-based multitask learning models have been used to relate sorbent and catalyst formulation variables to multiple performance metrics \cite{nkulikiyinka2022prediction}. Response-surface models have also proven effective for identifying promising oxide mixtures for oxidative coupling of methane, as demonstrated in a formulation-focused study by Nishimura et al. \cite{nishimura2022machine}, and for tuning the composition and preparation parameters of Cu-CHA catalysts through statistically guided parameter mapping \cite{ohyama2022bayesian}. In addition, Alanazi et al.\ used response-surface models to optimize Ni–zeolite catalysts for methane conversion \cite{alanazi2024response}, and Zarei-Jelyani et al.\ employed response-surface models to tune metal ratios and promoter levels in Ni–Co–Ce/Al\textsubscript{2}O\textsubscript{3} catalysts for steam-methane reforming \cite{zareijelyani2023response}. These studies illustrate how response-surface modeling can efficiently map catalyst formulation variables to activity and selectivity, acting as a form of experimental design that reduces the number of costly synthesis–testing cycles.

Generative AI techniques are opening a new frontier in catalyst discovery by enabling models to propose novel compositions and structures directly, rather than merely screening pre-existing candidates. Recent work on high-entropy alloy catalysts has shown that variational autoencoders (VAEs) can learn latent representations of multicomponent compositions and then generate previously unexplored alloy formulations with predicted catalytic promise \cite{lin2025llmdriven}, while condition-aware models such as CatDRX extend this idea by conditioning a VAE on reactants and products to generate homogeneous catalysts tailored to specific reaction environments \cite{kengkanna2025catdrx}.

Large-language-model (LLM) approaches offer a complementary generative strategy that takes advantage of natural language interfaces and pre-trained models. CatGPT demonstrates that transformer models can generate plausible catalyst formulations directly from textual prompts \cite{mok2024catgpt}, and CataLM \cite{wang2024catalm} and related LLM-based efforts \cite{caldasramos2023bayesian} show how domain-specialized language models can suggest catalyst compositions, synthetic precursors, and design rationales grounded in chemical patterns learned from the training data or in-context prompts. More broadly, general frameworks such as MatterGen provide a route for sampling chemically valid crystal structures across large materials spaces, including catalytic materials, providing a route to establish new candidate materials \cite{zeni2025generativemodel}. Finally, explainable generative models provide a degree of interpretability by linking latent-space directions to chemically meaningful formulation features, helping guide the rational modification of generated catalyst candidates \cite{semnani2024machinelearning}. Together, these methods illustrate how generative AI is shifting catalyst discovery from screening toward creative, model-driven exploration of vast compositional and structural spaces.

Recent advances in autonomous experimentation are beginning to reshape catalyst discovery by merging high-throughput robotics with machine learning in fully closed-loop ``smart lab'' workflows. For example, Bennett et al.\cite{bennett2024autonomous} demonstrated the design of homogeneous catalysts for olefin hydroformylation using an autonomous ``robotic chemist'' that independently designed, synthesized, and optimized the ligands of a new Rh-based catalyst system while performing hundreds of experiments per day in a flow chemistry platform. Zhang et al.\cite{zhang2025multimodal} extended this concept to electrocatalysis by using a multimodal robotic platform with integrated rapid ink preparation, automated electrode fabrication, and active-learning–guided electrochemical testing to identify a top-performing multi-element catalyst for formate oxidation within only a few iterative loops. Bayesian-optimization–driven discovery has also been used for identification of new catalyst formulations for the reverse water–gas shift reaction \cite{wang2023accelerated}, synthesis of higher alcohols from syngas \cite{suvarna2024activelearning}, and the OER reaction in acid \cite{jenewein2024navigating}. Robotics-integrated platforms have further expanded electrocatalyst discovery by combining automated synthesis with ML-guided selection of experiments \cite{kodera2023robotic} and enabled closed-loop identification of chiral Rh-ligand homogeneous catalysts through autonomous enantioselectivity screening \cite{kalikadien2024asymmetric}. These studies demonstrate how connecting AI and ML tools with automated experiments in fully or partially closed-loop systems can accelerate the discovery of new catalyst compositions.

Table \ref{tab:summary-kinetics} summarizes the findings of this section.

\begin{longtable}{@{}p{0.30\linewidth}p{0.32\linewidth}p{0.32\linewidth}@{}}
\caption{Kinetics, Catalysis and Reaction Engineering: current capabilities, remaining barriers, and future opportunities.}\label{tab:summary-kinetics}\\
\toprule
\textbf{Current capabilities} & \textbf{Remaining barriers} & \textbf{Future opportunities} \\
\midrule
\endfirsthead
\multicolumn{3}{l}{\textit{(continued)}}\\
\toprule
\textbf{Current capabilities} & \textbf{Remaining barriers} & \textbf{Future opportunities} \\
\midrule
\endhead
\bottomrule
\endfoot
\bottomrule
\endlastfoot
MLIPs reproduce DFT-level adsorption energetics ($\sim$0.2--0.5 eV) at orders-of-magnitude lower cost, enabling active-site, nanoparticle- and micron-scale surface simulations & MLIP accuracy is bounded by the underlying DFT training data; long-timescale trajectories are difficult to validate, and UQ outside the training domain remains immature & Tighter active-learning loops coupling MLIPs with DFT, and systematic UQ benchmarks for long-timescale/ out-of-domain predictions \\
AI-based mechanism generators (e.g., RMG-Cat, Genesys-Cat) and diffusion-based adsorbate-placement models automate construction of catalytic reaction networks & Estimating rate coefficients for every elementary step, adsorbate configuration, and facet remains combinatorially difficult; many ML rate models are not mechanistically interpretable & Broader use of physics-constrained (mass-action, site-conservation, thermodynamically consistent) kinetic ML to reduce reliance on empirical rate rules \\
ML surrogates for coverage-dependent kinetics and microkinetic/reactor coupling (e.g., lattice CNNs, AMUSE) enable reactor-scale simulation with retained mechanistic fidelity & Validation of learned kinetic models still relies heavily on comparison to DFT rather than synthesis and testing of real catalysts & Extending multiscale, DFT-to-reactor automation frameworks to routine industrial reactor design workflows \\
ML directly extracts rate coefficients and full rate laws from reactor/experimental data (SINDy-type and kinetics-informed neural networks), and LLM-assisted tools (e.g., SKAI) translate reaction descriptions into model equations & Data quality/quantity requirements for reliable kinetic discovery are often unmet in typical experimental datasets & Development of shared kinetics data benchmarks (e.g., Catechol, Summit, ORDERLY) to enable reproducible comparison across methods \\
High-throughput ML screening, generative (VAE, diffusion) and LLM-based (CatGPT, CataLM) models, and autonomous ``self-driving lab'' platforms accelerate catalyst discovery and optimization & Generated catalyst candidates are not always synthetically accessible; autonomous platforms remain limited to specific reaction classes and equipment configurations & Closer integration of generative catalyst design with autonomous synthesis-testing loops and multiobjective (activity/selectivity/sustainability) optimization \\
Emerging kinetics data benchmarks (e.g., Catechol, Summit, ORDERLY) provide shared datasets for reaction-kinetics ML & Code, trained MLIP/kinetic-model parameters, hyperparameters, and data splits are rarely released alongside published results, making it difficult to reproduce reported rate coefficients or reimplement mechanism-generation pipelines & Journal- and community-level requirements for code, model weights, and versions of software/environments as a condition of publication, mirroring practice in adjacent computational chemistry venues \\
\end{longtable}

\section{Thermodynamics, Transport and Fluid Mechanics}
\subsection{Transport Phenomena and Fluid Mechanics}

Recent advances in AI/ML have broadened their applicability in transport phenomena and fluid mechanics, extending from data analysis to predictive modeling, design, and control across a wide range of flow systems. The studies discussed in this section share a common objective of using AI/ML to extract useful physical insights from complex experimental and computational data, improve the analysis of multiscale transport processes, and enhance predictions of nonlinear, multiphase, and three-dimensional unsteady flows. Although the applications span droplet-laden flows, bubble columns, boiling and phase-change systems, microfluidics, and environmental flows, they collectively illustrate the growing convergence of data-driven methods with classical transport theory. This progression has led to the development of hybrid and physics-based frameworks that combine machine learning with first-principles models to provide accurate, interpretable, and physically consistent predictions.

\subsubsection{Analysis of complex multiphase phenomena}

Machine learning has emerged as a transformative framework in fluid mechanics, enabling the analysis, prediction, and control of complex multiphase phenomena that are often difficult to capture using traditional theoretical or experimental methods \cite{del2026perspective, taira2025critical, brunton2020machine}. In multiphase systems, such as droplet-laden flows, processes including droplet deformation, fragmentation, coalescence, evaporation, and transport are inherently nonlinear, multiscale, and highly sensitive to small variations in operating conditions \cite{chakraborty2026evaporation, ade2024droplet}. Conventional design and optimization strategies, relying on theoretical models, empirical correlations, or iterative experimentation, often struggle to efficiently resolve the associated spatiotemporal complexity \cite{gao2013uncertainty, shao2020machine, auyeung2023predicting}. Recent progress in high-speed imaging, digital in-line holography, tomographic diagnostics, and high-fidelity simulations has enabled ML models to extract droplet characteristics and underlying flow physics with unprecedented accuracy, supporting both fundamental research and industrial applications \cite{deepak2025droplet, ade2024application, ade2023swirl, ade2023breakup, kiyani2024partial}. A notable recent development is the integration of deep learning architectures for droplet segmentation and flow-field reconstruction from computationally expensive simulations and complex experimental image datasets \cite{katre2021evaporation}. Neural network architectures such as U-Net, R2-UNet, attention-based frameworks, and residual models can identify droplets directly from raw holographic or shadowgraphic data without conventional feature engineering \cite{barbastathis2019computational, guildenbecher2017fragmentation}. Once trained, these models can rapidly process thousands of images, enabling time-resolved quantification of droplet size distributions (DSDs) across diverse fragmentation regimes, including bag breakup, sheet thinning, and ligament breakup. Compared with conventional thresholding or Hough-transform-based methods, ML-driven segmentation is particularly effective when droplets are overlapping, highly deformed, or form dense sprays \cite{ade2024application}. Beyond droplet detection, regression and classification models enable direct prediction of droplet size, velocity, and breakup mode from experimental or simulation data \cite{talebjedi2022microdroplets}. Furthermore, convolutional neural networks (CNNs), recurrent neural networks (RNNs), and physics-informed neural networks (PINNs) are now widely applied to characterize and model turbulent flows \cite{hanrahan2023turbulence}. These capabilities increasingly support inverse-design workflows, including atomizer optimization, prediction of raindrop formation in atmospheric flows \cite{chakraborty2025droplets, takeishi2024parameterizing}, and improvements in spray-based combustion systems \cite{ade2024application, ade2023swirl, ade2023breakup}.

AI/ML approaches have become indispensable tools for analyzing multiphase flows in industrial and environmental systems \cite{chakraborty2025droplets, takeishi2024parameterizing, zhu2022review}. In bubble column reactors, widely used in chemical, petrochemical, and biochemical industries, ML models have been employed to track bubble breakup, coalescence, residence time, and interfacial area, all of which critically influence mass transfer and reaction rates. Deep learning frameworks further enable reconstruction of bubble shapes from experimental or simulated data and prediction of gas holdup across a range of operating conditions \cite{jiang2024bubble, zhu2022review}. These capabilities support accurate modeling of mixing, turbulence modulation, and transitions between flow regimes such as homogeneous, heterogeneous, and churn-turbulent states. Given the highly unsteady, three-dimensional nature of flow in bubble columns and the strong interphase coupling of heat and mass transfer, ML provides an efficient and scalable alternative to computationally expensive high-fidelity simulations \cite{zhao2024mlcfd}.

Similar advances have been demonstrated in liquid–liquid extraction units, slurry reactors, fluidized beds, coastal sediment transport, and thermal phase-change systems, where ML facilitates real-time detection and quantification of dispersed phases \cite{wang2026bubble}. For example, in boiling heat transfer, segmentation networks accurately identify nucleation sites and departure diameters, while sequence-based models capture the temporal evolution of bubble dynamics. In spray cooling, AI models can monitor droplet size, velocity, and surface coverage to dynamically adjust nozzle flow rates or heating profiles for optimal thermal performance. In biomedical microfluidics, ML-controlled droplet generation enables adaptive modulation of breakup and merging processes for applications such as on-chip chemical reactions or single-cell encapsulation. In atmospheric and environmental applications, ML-enabled characterization of droplets and aerosols has improved modeling of rainfall formation, cloud microphysics, and pollutant transport. In particular, ML frameworks now predict droplet size distributions and inform strategies for cloud seeding, aerosol injection, and weather modification \cite{chakraborty2025droplets, takeishi2024parameterizing}. By integrating high-resolution measurements with physics-informed ML models, researchers are able to optimize droplet formation, growth, and coalescence under varying thermodynamic conditions, improving accuracy of rainfall predictions and supporting targeted environmental interventions.

These advances reduce reliance on large-scale field measurements while offering actionable guidance for engineering design and policy decisions. The convergence of ML with advanced sensing technologies, such as high-speed imaging, holography, and tomographic diagnostics, now enables real-time optimization, where fluid systems can adapt dynamically based on measured or predicted performance.

\subsubsection{Hybrid modeling: Interfacing ML and physics-based models}
\label{sec:hybrid-modeling-interfacing-ml-and-physics-based-models}
One promising frontier is the synergy between ML and physics-based modeling \cite{hanrahan2023turbulence, qiu2025pinn}. Hybrid approaches, incorporating physical constraints into neural networks or using ML to accelerate Computational Fluid Dynamics (CFD) solvers, are reshaping droplet-laden flow simulations. PINNs, for example, solve governing equations like Navier–Stokes directly from data, reconstructing velocity and pressure fields from sparse measurements. For droplet dynamics, these models predict breakup thresholds, surface instabilities, and interfacial forces without requiring full numerical simulations. Reinforcement learning (RL) is also being explored for multiphase system control, such as optimizing atomization processes, modulating bubble distributions in reactors, and designing active feedback in spray cooling. With advances in sensor networks and high-speed imaging, real-time ML-based control of multiphase flows is becoming increasingly feasible. In industrial practice, AI/ML-driven design and optimization demonstrate clear advantages across chemical, petrochemical, and process engineering applications. For example, in bubble column reactors, models trained on experimental or CFD data predict the effects of operational changes on bubble breakup, coalescence, residence time, and interfacial area, enabling optimization of reactor geometry, sparger design, and flow conditions without extensive experimental campaigns \cite{jiang2024bubble, zhu2022review, zhao2024mlcfd}. In heat-transfer systems, ML models optimize boiling bubble dynamics or fine-tune droplet spray cooling patterns for enhanced thermal performance. In liquid–liquid extraction units, ML guides impeller and packing design and flow rates to improve separation efficiency by predicting droplet dispersion and coalescence under varying operational conditions \cite{wang2026bubble}.

Looking ahead, the integration of AI/ML into fluid mechanics represents a paradigm shift in system design and optimization. While traditional analytical, empirical, and computational approaches remain foundational, AI/ML complements them by enabling rapid, high-dimensional optimization, real-time feedback control, and exploration of otherwise intractable design spaces. The combination of physics-informed modeling, deep learning-based data extraction, and reinforcement learning-driven control is creating fluid systems that are self-optimizing, adaptive, and highly efficient.

 Overall, AI and ML are evolving from tools for analysis or simulation to central components of modern fluid system design. By integrating high-fidelity measurement, predictive modelling, and optimization within a unified framework, these approaches reduce development cycles, lower costs, enhance operational efficiency, and enable next-generation devices across industrial, environmental, and biomedical applications. From high-performance chemical reactors to advanced droplet-based microfluidics and large-scale atmospheric interventions, AI/ML-driven design is transforming how fluid systems are conceived, optimized, and controlled, ushering in a new era where intelligent, data-driven strategies are indispensable to fluid mechanics research and engineering \cite{ade2024application, zhu2022review, qiu2025pinn}.


The models discussed in this section directly address several of the key challenges outlined in the Introduction. In terms of predictive accuracy, ML-enabled frameworks leverage high-fidelity simulation and experimental data to capture complex, nonlinear multiphase dynamics that are difficult to resolve using conventional models. Challenges associated with limited or noisy data are mitigated through hybrid and physics-informed approaches, which incorporate governing equations and known physical constraints to regularize learning and improve robustness. Importantly, many of these models explicitly enforce material and energy conservation by embedding conservation laws (e.g., mass, momentum, and energy balances) into the model architecture or loss functions, as in physics-informed neural networks and hybrid CFD–ML frameworks. This ensures physically consistent predictions, even in regimes with sparse data or during extrapolation. Furthermore, interpretability is enhanced by integrating ML components within established transport equations, enabling model predictions to be analyzed in the context of known fluid-mechanical principles. Finally, these approaches improve extrapolation capability and model reliability by restricting predictions to physically admissible solution spaces, thereby aligning data-driven learning with first-principles understanding.

\subsection{Thermodynamics}


\subsubsection{Thermodynamic and transport property prediction}

Over the past decade, the number of publications explicitly incorporating ML for thermodynamic property prediction has grown remarkably, with a large fraction relying on regression algorithms and artificial neural networks (ANNs) to represent thermodynamic behavior, especially for systems where traditional physical models are difficult to develop or for emerging families of compounds that can be tuned and easily expanded, such as ionic liquids or deep eutectic solvents \cite{tian2025intelligent}. While many contributions have reported successful results for specific properties, a central limitation of supervised learning methods is the need for large and well-labeled datasets, whereas thermophysical property data are often incomplete or unevenly reported. Consequently, many ANN-based studies have been hindered by data scarcity, inconsistency, and limited coverage of chemical space \cite{jirasek2021perspective}.

Additionally, ML has also been increasingly used to estimate transport properties such as viscosity\cite{bilodeau2023machine,braun2025interpretable}, thermal conductivity\cite{barua2024interpretable,braun2025interpretable}, ionic conductivity\cite{ma2026machine}, and diffusion coefficients\cite{dias2024machine}---properties that lie beyond the predictive scope of most equations of state and typically rely on semi-empirical correlations. Data-driven models can offer competitive or superior predictive performance for these properties by leveraging molecular descriptors, state variables, and nonlinear regression algorithms. This body of work has also highlighted the importance of model interpretability and feature relevance when applying ML to transport phenomena, emphasizing both the opportunities and methodological challenges associated with replacing or complementing traditional correlations \cite{slepavicius2023application,chaparro2024simulation}.

Another key trend in recent years is the incorporation of thermodynamic constraints directly into the ML architecture or loss function. These ``thermodynamics-informed'' ML approaches embed identities such as the Gibbs--Helmholtz or Clausius--Clapeyron relations, ensuring internal consistency and improving extrapolation performance in low-data regimes\cite{hoffmann2026thermodynamically}. Integrating molecular information with explicit thermodynamic constraints improves robustness, reduces unphysical predictions, and relates ML outputs to established physical behavior \cite{hammad2025multioutput}. This effort aligns with the broader shift toward ML models that respect conservation laws, convexity of free energies, stability of phases, and other fundamental requirements of thermodynamics, as discussed in Section \ref{sec:hybrid-modeling-interfacing-ml-and-physics-based-models}.

As also emphasized in Section \ref{sec:simulations-of-catalysts}, at the microscopic level, ML has transformed atomistic simulations, which are central to thermodynamic property prediction. State-of-the-art machine-learned interatomic potentials can now reproduce \emph{ab initio} accuracy at a fraction of the computational cost, making it feasible to evaluate thermodynamic properties of large systems over long timescales \cite{kulichenko2024datageneration}. Simultaneously, ML within statistical mechanics has enabled data-driven mappings between microscopic configurations and macroscopic thermodynamic functions, accelerating the estimation of free energies, phase behavior, and phase diagrams, even expanding exploration of compositional and thermodynamic state spaces \cite{xu2025nepmbpol}.

Parallelling recent trends in catalyst design (Section \ref{sec:Accelerating-high-throughput-atomic-scale-computational-screening}), beyond forward prediction, ML is increasingly used for inverse problems: given target thermodynamic behavior, infer molecular architecture, mixture composition, or operating conditions. Several recent efforts combine chemical representations with SAFT-based approaches to map SMILES strings and molecular fingerprints \cite{winter2025predicting, habicht2023predicting}, or $\sigma$-profiles \cite{alkhatib2022searching} to equation of state (EoS) parameters, enabling rapid screening of virtual molecular libraries for desired phase behavior or critical properties.

 A complementary line of work replaces or augments the EoS with neural networks constrained by thermodynamics. Neural-network surrogate models can be trained to represent free energies, from which all thermodynamic quantities are derived, thereby ensuring consistency by construction \cite{qu2024nneosc}. Similar models explicitly embed phase transitions and stability criteria within the network architecture and loss function \cite{felton2024mlsaft}. Collectively, these methods mark a shift from purely black-box regressors toward hybrid frameworks in which ML augments, rather than replaces, thermodynamic theory.

Despite impressive progress, multiple challenges remain. Data scarcity, inconsistency, and heterogeneity across sources (different laboratories, measurement standards, or purity levels) limit the reliability of purely data-driven models. Ensuring thermodynamic consistency, respecting Maxwell relations, convexity of free energies, and positivity of entropy production remains difficult for generic ML architectures, motivating continued development of physics-informed designs. Extrapolation beyond the training domain, crucial for process synthesis and safety, remains a persistent weakness. Furthermore, integrating ML-based thermodynamic modules into  process modeling/design software  requires robust uncertainty quantification, transparent validation, and long-term maintainability.

 Overall, the field is evolving from simple property regression toward hybrid frameworks in which ML augments, rather than replaces, thermodynamic modeling. The combination of high-fidelity simulation data, curated experimental databases, and physics-informed ML architectures promises fast, accurate, and generalizable thermodynamic models for emerging fluids and applications. Table \ref{tab:summary-thermo} summarizes the findings of this section.

\begin{longtable}{@{}p{0.30\linewidth}p{0.32\linewidth}p{0.32\linewidth}@{}}
\caption{Thermodynamics, Transport and Fluid Mechanics: current capabilities, remaining barriers, and future opportunities.}\label{tab:summary-thermo}\\
\toprule
\textbf{Current capabilities} & \textbf{Remaining barriers} & \textbf{Future opportunities} \\
\midrule
\endfirsthead
\multicolumn{3}{l}{\textit{(continued)}}\\
\toprule
\textbf{Current capabilities} & \textbf{Remaining barriers} & \textbf{Future opportunities} \\
\midrule
\endhead
\bottomrule
\endfoot
\bottomrule
\endlastfoot
Deep learning (CNNs, U-Net, attention/residual architectures) segments droplets/bubbles and reconstructs multiphase flow fields directly from holographic and simulation data, without hand-engineered features & Conventional design/optimization strategies (empirical correlations, iterative experiments) still struggle with the spatiotemporal complexity of multiphase flows, and purely data-driven models can violate physical constraints & Continued convergence of high-speed/holographic sensing with physics-informed ML for real-time, self-optimizing multiphase systems \\
Hybrid and physics-informed frameworks (e.g., PINNs) embed mass/momentum/energy conservation directly into architectures or loss functions, improving robustness under sparse or noisy data & Real-time ML-based control of multiphase flows is still constrained by sensing bandwidth and computational cost & Broader deployment of RL-based control for atomization, bubble-column operation, and spray cooling in industrial settings \\
ML/ANN models predict thermophysical and transport properties (viscosity, thermal and ionic conductivity, diffusion coefficients) competitively with or better than semi-empirical correlations, especially for tunable compound families (e.g., ionic liquids) & Thermophysical/transport property data remain scarce, inconsistent, and heterogeneous across sources, limiting the reliability of purely data-driven models & Curation of standardized, cross-laboratory thermophysical property databases to reduce data heterogeneity \\
Thermodynamics-informed ML (embedding Gibbs--Helmholtz, Clausius--Clapeyron and related identities) and ML-augmented/neural EoS improve internal consistency and extrapolation in low-data regimes & Enforcing full thermodynamic consistency (Maxwell relations, convexity of free energies, positivity of entropy production) remains difficult for generic architectures & Deeper architectural embedding of stability/phase-equilibrium constraints so ML augments, rather than replaces, EoS-based thermodynamic theory \\
Inverse-design workflows map molecular representations (SMILES, fingerprints, $\sigma$-profiles) to EoS parameters, enabling rapid virtual screening for target phase behavior & Extrapolation beyond the training domain, critical for process synthesis and safety, remains a persistent weakness; integration into commercial process software  requires robust UQ and validation & Development of transparent, well-validated ML thermodynamic modules with long-term maintainability for integration into standard process simulators \\
Neural EoS surrogates and thermodynamics-informed property-prediction models are increasingly reported in the literature & Trained network weights, training/test splits, and the specific software/library versions used to fit thermodynamic surrogates are seldom shared, hindering independent verification against reported accuracies & Minimum code- and data-availability requirements for published thermodynamic/transport ML models, enabling direct reproduction and fair cross-study comparison \\
\end{longtable}

\section{Separations}

ML has had significant impacts on separation processes across length scales, from materials to process design and optimization.



\subsection{Gas-Liquid Contact Operations}
Distillation remains the key technology for separating liquid mixtures in industry \cite{jirasek2025mccabe}. The fundamentals principles are generally well understood, but applications challenges remain, particularly in the case of separating complex and/or multi-component streams. These include, e.g., mixtures encountered in refining, that are characterized using empirical fractions rather than at the molecular level \cite{colling2025data,long2025neural}, and mixtures with impurities \cite{eldi2025enhancing}.

At the design level, \citet{montesinos-castellanos_2026} demonstrated the potential of generative AI for distillation column design. Leveraging variational autoencoders, it was shown that generative AI can generate novel configurations, ensure purity constraints, and boost efficiency, while outperforming discriminative AI in process intensification innovation. \citet{park2022framework} demonstrated the use of ML for optimizing the energy use in distillation, while \citet{mohd2023machine} undertook a similar effort at the process level.

\citet{lu2025interpretable} discuss the use of interpretable ML for predicting the performance of high-gravity NOx oxidation/absorption, a complex intensified operation. In a similar vein, \citet{saraswathi2020random} used ML tools to predict hydrodynamic properties in a rotating disc contactor.

\subsection{Separation Materials}
\label{sec:separation-materials}

\subsubsection{Adsorbents}


Among the emerging classes of adsorbents, metal–organic frameworks (MOFs) have been by far the most widely studied due to their high surface areas and tunable adsorption properties \cite{lee2021computational,wilmer2012largescale,rosen2021machine}. Their reticular structure and thus modular nature also make them an excellent candidate for ML-based study. One of the most significant enabling studies in this field was the release of the Computation-Ready Experimental (CoRE) MOF database by Snurr and colleagues.\cite{chung2014computation} This database and subsequent updates and extensions have provided researchers with tens of thousands of structural files of MOFs that have been experimentally synthesized and characterized in wet labs across the world.\cite{zhao2025core} This database has provided a fertile ground for using ML to explore the gas separation properties of MOFs for applications such as noble gas separation,\cite{zhao2023highthroughput} paraffin/olefin separation,\cite{tang2021rapid,xue2023highthroughput} and CO$_2$ capture.\cite{orhan2023accelerating,zhang2021machine,choudhary2022graph} These workflows have even been integrated with Ideal Adsorbed Solution Theory calculations to predict binary gas adsorption behavior.\cite{yu2023efficient}

Several studies have leveraged ML to navigate this expanded MOF space consisting of millions of adsorbent candidates.\cite{lee2021computational,cao2023moformer,kang2023multimodal} Of note, the use of ML has now moved far beyond simple screening. For example, one of the most exciting applications is the use of ML models to generate entirely new MOFs for targeted applications.\cite{lu2025generative} Fu et al.\ developed MOFDiff, which leveraged a diffusion model architecture to create coarse-grained MOF representations that were then decoded to full atomistic representation and optimized for CO$_2$ capture.\cite{fu2023mofdiff} In another study, Aspuru-Guzik and co-workers used a variational autoencoder architecture to generate a customized text-based representation that was then used to build all new MOFs for CO$_2$ separations, resulting in several candidates competitive with the state of the art.\cite{yao2021inversedesign} Generative MOF design is now incorporated into large language model (LLM) studies through the development of ChatMOF by Kang and Kim. They leveraged GPT-4 to create an agentic pipeline for searching, predicting, and generating new MOFs targeted at CO$_2$ capture.\cite{kang2024chatmof} Clearly, ML has opened a world of exciting possibilities to translate materials discovery with pressing gas separations challenges that are critical to chemical engineering.

\subsubsection{Membranes}
With the rapid development of membrane separation technologies based on emerging functional materials --- including two-dimensional materials \cite{xu2013graphenelike,tan2017recent,chakraborty2021twodimensional} (e.g., graphene, graphene oxide and MXene), framework materials (e.g., metal--organic frameworks (MOFs) \cite{qian2020mof} and covalent organic frameworks (COFs) \cite{parkinson2025toward}), and advanced porous polymers \cite{evans2022twodimensional} --- it has become possible to achieve highly selective separation of small ions and molecules for applications ranging from water desalination to resource recovery and gas purification. This progress stems from the rational engineering of membrane structures, surface chemistries, and pore architectures to precisely regulate transport pathways. For example, in water treatment processes, multiscale separations occur through a combination of microfiltration, ultrafiltration, nanofiltration, and reverse osmosis, requiring fine control of molecular sieving and electrostatic exclusion mechanisms \cite{liao2018progress}. Similarly, lithium extraction from brines demands sub-nanometer discrimination between Li$^+$ and Mg$^{2+}$ ions \cite{zhang2020membrane,hou2021lithium}, while gas separation membranes often rely on subtle differences in molecular diffusivity and adsorption enthalpy to achieve efficient CO$_2$/N$_2$ or H$_2$/CO$_2$ separation \cite{qian2020mof,qu2022pore}. In this context, the integration of AI into separation processes is opening new avenues for understanding and optimizing selective transport phenomena, bridging macroscopic flow behavior with angstrom-scale interfacial interactions within confined pores.

ML has  been extensively used in the design and development of new materials for polymeric membranes. In particular, Ramprasad and colleagues have contributed to the development of the Polymer Genome database. At release, Polymer Genome uses a fingerprinting scheme to featurize polymers for surrogate models and predict a variety of useful physical properties such as glass transition temperature, dielectric constant, etc.\cite{kim2018polymer} It has since undergone several updates to include prediction of gas permeability through membranes formed from polymers.\cite{zhu2020polymer} Ramprasad and colleagues have also made further contributions to the field by developing multi-task ML models that can predict diffusion coefficients, solubility coefficients, and permeabilities for both gases\cite{phan2024gas} and complex organic solvent separations,\cite{nistane2025polymer} highlighting both the diversity of potential use cases of polymer membranes in chemical separations and the utility of ML in navigating the structure–performance space in a rational and physically insightful way.

ML has also been applied to other topics within membrane science beyond gas separation performance with pure polymers. For example, mixed-matrix membranes (MMMs), in which secondary materials are dispersed in polymers to improve overall transport properties, are widely studied. ML has been used for \textit{in silico} screening to find the optimum dispersed/continuous phase combinations for industrially relevant separations such as CO$_2$/CH$_4$\cite{guan2022design} and H$_2$/CH$_4$.\cite{daglar2022combining} Another interesting application is using ML to understand the most relevant factors to ion transport through nanoporous membranes, which has implications in both water treatment and metal recovery.\cite{ritt2022machine} A recurring trend in many of these studies is the use of experimental data and its homogenization with selected data derived from molecular simulations, which stands in contrast to most ML studies of MOF adsorbents that use entirely simulated datasets from molecular dynamics or Grand Canonical Monte Carlo methods. This observation supports that quality ML models can be developed with thoughtful design, even on comparably smaller datasets, which highlights the benefit of using data from disparate sources to build more robust models.

\subsection{Module and Process Modeling}
In addition to the vast body of work on using ML to improve the molecular design of separations materials, it has also been used by several researchers to improve the design of process units using these materials. Adsorption processes are a common target due to their well-established operating principles and well-defined governing equations.

Pai et al.\ developed the machine-assisted adsorption process learning and emulation (MAPLE) framework, built on a dense feedforward network with Bayesian regularization, to optimize a vacuum swing adsorption (VSA) cycle for CO$_2$/N$_2$ separation from single-component isotherms.\cite{pai2020generalized} They later built on this model by incorporating governing differential transport equations and expanding the range of application to include chromatography to create a physics-based artificial neural network framework for adsorption and chromatography emulation (PANACHE). PANACHE achieved 100\,$\times$ improvement in computation time compared to traditional solvers while maintaining a high degree of fidelity to the underlying physics.\cite{subraveti2022physicsbased}

\subsubsection{Multiscale Modeling and Design}
\label{sec:multiscale-modeling-and-design}
Thermodynamics defines the energy landscape governing selective transport across membranes, linking driving forces such as osmotic pressure, chemical potential gradients, and interfacial free energies with separation efficiency.

Conventional, physics-based modeling platforms that are frequently used in this field, such as MATLAB and COMSOL rely on deterministic formulations (e.g., Poisson--Nernst--Planck equations \cite{cervera2005poisson,vlassiouk2008ionic}) to describe ion diffusion, advection, and permeation through selective media.

However, these approaches often struggle to capture the nonlinear couplings among pore structure, surface chemistry, and fluid behavior. In subnanofluidic and ion-selective systems, traditional thermodynamic models also fail to account for multicomponent interactions, hydration-shell restructuring, and specific surface--solute interactions that govern selectivity. Although molecular dynamics simulations can quantify energy barriers, binding energies, and dehydration effects for ions traversing subnanometer channels \cite{ban2012molecular,abraham2017tunable,lu2020efficient}, they remain limited by computational cost and system size.

As in the case of kinetics and thermodynamics modeling (Sections \ref{sec:Accelerating-high-throughput-atomic-scale-computational-screening} and \ref{sec:hybrid-modeling-interfacing-ml-and-physics-based-models}, AI provides a powerful complementary approach by learning membrane-relevant thermodynamic relationships directly from experimental and simulation datasets while embedding physical constraints \cite{ibnouf2025comprehensive,gao2023understanding,elhakim2025ai,jafari2024ai,thajudeen2024integration}. Machine learning models can accurately predict key thermodynamic parameters---including osmotic pressures, activity coefficients, and hydration enthalpies---across wide ionic-strength and mixed-electrolyte ranges, frequently outperforming empirical correlations \cite{raven2024machine,liang2025machine,zamfir2025application,cao2024machine}.

Hybrid thermodynamic--AI frameworks further enable dynamic coupling of energetic terms (chemical potential differences, interfacial free-energy barriers, entropy changes) with transport kinetics, yielding more predictive descriptions of membrane processes and allowing real-time optimization under variable operating conditions. By integrating high-dimensional data from MD, transport measurements, and process operation with fundamental physical laws, AI can uncover hidden transport mechanisms, guide membrane design, and enhance predictive performance across applications ranging from high-salinity desalination to mixed-gas separations.


In the realm of adsorption, Grimm and Gazzani relied on multiscale modeling for selecting optimal sorbents for dilute CO$_2$ capture via a vacuum–temperature swing adsorption process. Their model was highly computation-efficient because they reduced their modeling framework to a series of simple algebraic equations rather than more complex partial differential equations and were able to rapidly screen over 2,500 potential adsorbents and validate their viability for a full-scale adsorption process.\cite{grimm2022machine} In another integrated study, Ji and co-workers developed a random forest ML model to predict the ethane/ethylene adsorption behavior of hundreds of thousands of MOFs and then used this information with a VSA model to identify several candidate MOFs with high process-level performance.\cite{cheng2025machine}  From these examples, it is clear that ML will be a critical tool to the future improvement of all aspects of  separation processes.

\subsubsection{System Operations}
Beyond predictive modeling, AI is also revolutionizing how fluid transport is controlled and engineered in separation systems. Reinforcement learning and advanced optimization algorithms can autonomously tune operating parameters such as pressure gradients, concentration distributions, and temperature fields to maximize selectivity and permeation flux while minimizing energy consumption. For example, in water desalination, AI-based control strategies can adaptively adjust operating conditions to minimize fouling and energy use \cite{ibnouf2025comprehensive}, whereas in lithium extraction, AI may optimize electro-driven transport pathways to enhance selectivity between competing ions \cite{zhang2020membrane,hou2021lithium}. Generative models and inverse design strategies further accelerate the discovery of novel membrane materials and channel architectures with targeted transport properties. Importantly, AI can identify emergent behaviors at complex fluid--solid interfaces --- including cooperative ion transport, nanoscale flow instabilities, and dynamic fouling phenomena --- that are often inaccessible to conventional analytical or numerical methods. This convergence of AI and transport science is poised to transform membrane separation engineering from an empirical discipline into an intelligent, adaptive, and predictive technology platform.

Table \ref{tab:summary-separations} summarizes existing progress and future opportunities in the Separations area.
\begin{longtable}{@{}p{0.30\linewidth}p{0.32\linewidth}p{0.32\linewidth}@{}}
\caption{Separations: current capabilities, remaining barriers, and future opportunities.}\label{tab:summary-separations}\\
\toprule
\textbf{Current capabilities} & \textbf{Remaining barriers} & \textbf{Future opportunities} \\
\midrule
\endfirsthead
\multicolumn{3}{l}{\textit{(continued)}}\\
\toprule
\textbf{Current capabilities} & \textbf{Remaining barriers} & \textbf{Future opportunities} \\
\midrule
\endhead
\bottomrule
\endfoot
\bottomrule
\endlastfoot
Generative AI (e.g., VAE-based) has been applied to distillation column design and energy-use optimization; interpretable ML predicts performance of intensified gas--liquid operations & Most ML research targets membranes and adsorption; distillation, still the dominant industrial technology, remains comparatively underserved by AI/ML advances & Extending generative and interpretable ML methods to established gas--liquid operations and complex/impure multicomponent streams \\
Large curated databases (CoRE MOF, Polymer Genome) and ML screening (multitask learning, random forest) enable rapid evaluation of MOF/polymer gas-separation and CO$_2$-capture performance across millions of candidates & MOF ML workflows rely largely on simulated (DFT/GCMC) data, whereas membrane ML studies mix experimental and simulated data, complicating cross-domain model transfer & Development of unified, experimentally grounded datasets spanning adsorbents and membranes to improve model transferability \\
Generative models (MOFDiff, VAE-based inverse design, ChatMOF) propose entirely new MOF and membrane candidates targeted at CO$_2$ capture and gas separation & Generated hypothetical adsorbents/membranes are not always synthetically accessible, and models trained on equilibrium isotherms struggle with dynamic, multicomponent industrial conditions & Coupling generative materials design with synthesis-feasibility filters and models that capture adsorption/desorption kinetics alongside thermodynamics \\
Physics-based ML process emulators (e.g., MAPLE, PANACHE) achieve $\sim$100$\times$ speedups over conventional solvers while retaining transport-equation fidelity for adsorption/chromatography cycles & Integration of molecular-scale material properties with process-level design and control remains only partially realized & End-to-end frameworks that translate material-level predictions directly into optimal process configurations \\
RL and optimization algorithms autonomously tune membrane and adsorption operating parameters (pressure, concentration, temperature) to improve selectivity and reduce energy use and fouling & Emergent, complex fluid--solid interfacial behaviors (cooperative ion transport, dynamic fouling) remain difficult to capture with conventional analytical or numerical methods & Broader real-time, AI-driven control of desalination, lithium-extraction, and gas-separation operations informed by AI-identified interfacial mechanisms \\
Process-level ML emulators (MAPLE, PANACHE) and generative adsorbent/membrane models are increasingly published with reported performance metrics & Underlying code, isotherm/simulation datasets, trained network parameters, and hyperparameters are inconsistently released, complicating independent verification of speedup and accuracy claims & Adoption of explicit code- and data-availability requirements for separations ML publications, comparable to those emerging in adjacent computational chemistry journals \\
\end{longtable}

\section{Materials and Interfaces, and Applied Chemistry}

The discovery and optimization of functional materials has long constrained the pace of technological advancement. Traditional experimental campaigns, from a pessimistic view reliant on serendipity and incremental refinement, struggle to navigate the combinatorial vastness of chemical composition space. As previewed in Sections \ref{sec:Accelerating-high-throughput-atomic-scale-computational-screening} and \ref{sec:separation-materials}, ML architectures now offer a fundamentally different paradigm. Rather than exhaustive enumeration, these computational frameworks learn underlying structure-property relationships from existing data, enabling rapid screening of candidate materials that would require decades of laboratory effort through conventional approaches \citep{butler2018machine}. This transformation extends beyond acceleration and reshapes how researchers can conceptualize the discovery process itself.


A comprehensive review of crystalline materials prediction highlights how ML models now routinely estimate solubility, melting point, adsorption capacity, band gap, and electrochemical properties with accuracy rivaling conventional density functional theory (DFT), while dramatically reducing computational cost \citep{chen_2024b}. Similarly, ensemble ML strategies combining XGBoost, SVM, ridge regression, and random forest models have been successfully applied to predict dielectric strength, dielectric constant, and electron affinity of organic compounds, demonstrating the maturity of supervised AI in high-dimensional chemical feature spaces \citep{shin_2024}. These studies collectively establish that data-driven models can capture complex nonlinear structure-property relationships across diverse chemical domains.

Graph neural networks (GNNs), which are notable tools for representing molecular and crystalline systems, have further advanced this capability by enabling descriptor-free learning directly from molecular graphs and periodic crystal structures. Instead of relying on predefined handcrafted features, GNN architectures encode atomic environments as nodes within connected graphs, capturing local coordination chemistry and long-range structural motifs simultaneously \citep{xie2018cgcnn}. The approach is flexible. Small organic molecules, extended inorganic frameworks, and amorphous interfaces can all be represented within unified computational schemas. This, in turn, facilitates learning across disparate chemical domains. \citet{chen2019graph} demonstrated that such graph-based representations achieve near-density functional theory (DFT) accuracy for a range of property predictions, including formation energies, band gaps, and elastic moduli, while reducing computational costs by several orders of magnitude. As discussed in recent reviews of crystalline property prediction, graph-based models excel in representing metal-organic frameworks (MOFs), energetic materials, and electrochemical systems across multiple scales \citep{chen_2024b}. These architectures allow unified treatment of small molecules, extended frameworks, and particle-scale crystalline products, facilitating cross-domain transferability and improved generalization.
Similar developments have been made in graph kernels, which can be used to build probabilistic models over molecules as graphs~\citep{xie2025bogrape, xu2021self}.

Importantly, AI-driven materials discovery has expanded beyond forward prediction toward generative and inverse design. Generative models such as variational autoencoders (VAEs) and conditional VAEs (CVAE) have been employed to explore vast chemical design spaces by learning continuous latent representations of molecular structures. For example, a CVAE framework was used to design supported ionic liquid membranes (SILMs) for CO\textsubscript{2} capture, generating candidate membranes from limited experimental data and achieving strong agreement between predicted and experimentally measured CO\textsubscript{2} absorption capacities \citep{ismail_2025}. In parallel, VAE models combined with graph neural network property predictors and reinforcement learning have enabled inverse molecular design of surfactants with targeted critical micelle concentration values, demonstrating controlled navigation of chemical latent space toward predefined property thresholds \citep{nnadili_2024}. These approaches mark a conceptual shift from predicting properties of known materials to proposing entirely new materials optimized for desired performance.

As emphasized in Section \ref{sec:Accelerating-high-throughput-atomic-scale-computational-screening}, that covered catalyst discovery, the broader frontier of materials informatics now lies in the integration of AI with laboratory automation. Recent advances demonstrate that the coupling of machine learning algorithms, including Transformers and large language models (LLMs), with high-throughput experimentation platforms has enabled autonomous optimization of MOF synthesis conditions \citep{zhao_2025}. The emergence of self-driving laboratories (SDLs), in which AI models iteratively propose experimental conditions, automated platforms execute synthesis and characterization, and newly acquired data refine the predictive model, represents a closed-loop paradigm for materials discovery. This convergence of predictive modeling, generative design, and robotic experimentation transitions AI from a passive analytical tool to an active driver of experimental decision-making.

Finally, an important distinguishing feature of many current-generation materials informatics models, relative to earlier quantitative structure-property relationship (QSPR) efforts, is the explicit encoding of physical constraints directly into the model architecture. E(3)-equivariant architectures \citep{batzner2022e3equivariant} build on group theory to ensure that learned representations transform correctly under rotations and translations. This mathematical system yields improvements in data efficiency, notably for models requiring thousands of training examples, which, under conventional architectures, can achieve comparable performance with hundreds when symmetry constraints are encoded. The implications for novel chemistries where data is scarce are significant.

\subsection{Benchmark Datasets and Computational Infrastructure}\label{benchmark-datasets-and-computational-infrastructure}

Standardized evaluation protocols underpin reproducibility in computational disciplines, and materials informatics is no exception. The Matbench initiative established systematic benchmarks for formation energy prediction, band gap estimation, and elastic modulus calculation, enabling comparisons across algorithmic approaches \citep{dunn2020matbench}. These datasets, curated from the Materials Project repository, encompass tens of thousands of compounds with DFT-computed properties, providing sufficient statistical power to distinguish genuine methodological advances from artifacts of specific train-test splits.

The Open Catalyst 2020 (OC20) dataset, designed to address the challenges of heterogeneous catalysis, represents perhaps the most ambitious community effort in this domain. It is comprised of over 1.2 million DFT relaxations across diverse surface-adsorbate configurations \citep{chanussot2021open}. Subsequent extensions through OC22 incorporated oxide electrocatalysts, addressing limitations of the initial release while expanding coverage to technologically critical reaction systems \citep{tran2023open}. Together, these resources contribute to the development of ML models capable of predicting adsorption energies with mean absolute errors approaching 0.2 eV. This is sufficient for preliminary catalyst screening, though still inadequate for quantitative kinetic modeling.


For large-scale materials screening, cloud high-performance computing (HPC) provides accessibility to the necessary computational resources. \citet{chen2024azure} demonstrated end-to-end workflows integrating ML geometric optimization with DFT validation, processing over 32 million candidate structures to identify 18 promising solid electrolytes for battery applications. The computational architecture leveraged Azure Quantum Elements, with ML inference requiring approximately 46 seconds per structure-relaxation step across 38,000 node-hours in total. Such throughput was unimaginable a decade ago. It reflects both algorithmic advances in ML potential accuracy and infrastructural investments in distributed computing platforms.

\subsection{Universal Machine Learning Potentials}\label{universal-machine-learning-potentials}

Continuing the discussion on MLIPs in Section \ref{sec:atomic-scale-simulations}, we note that the development of interatomic potentials applicable across the entire periodic table represents a key moment for computational materials science. Classical force fields, parameterized for specific element combinations, cannot capture the diversity of bonding environments encountered when exploring novel compositions. Universal ML potentials aim to circumvent this limitation by leveraging learned representations that generalize across the chemical space \citep{chen2022m3gnet}. M3GNet, trained on Materials Project relaxation trajectories, has been shown to achieve formation energy errors below 30 meV/atom while predicting forces with accuracy sufficient for geometry optimization. CHGNet extends this framework by incorporating explicit charge representations, thereby improving performance for ionic systems where  electrostatic interactions dominate \citep{deng2023chgnet}.

The GNoME project scaled these approaches dramatically. \citet{merchant2023gnome} reported the discovery of 2.2 million stable crystal structures through iterative cycles of ML prediction and DFT validation, expanding the known stable materials database by an order of magnitude. The significance extends beyond the sheer numbers. Many predicted structures occupy previously unexplored regions of composition space, suggesting that conventional intuition about chemical stability may be systematically incomplete. Through experimentation, novel phases have been validated, demonstrating that computational predictions can translate into laboratory reality with this approach.

For state-of-the-art accuracy with limited training data, equivariance constraints are essential. Tensor field networks introduced explicit rotational and translational symmetries via spherical harmonic expansions \citep{thomas2018tensor}, an approach further refined in the NequIP and Allegro architectures \cite{batzner2022e3equivariant}. These models achieve chemical accuracy on molecular dynamics benchmarks while maintaining computational efficiency compatible with nanosecond-scale simulations. The practical consequence is a step-change as researchers can now explore conformational landscapes and reaction pathways that previously required (prohibitively) expensive \textit{ab initio } molecular dynamics.

\subsection{Autonomous Discovery Platforms}\label{autonomous-discovery-platforms}

Laboratory automation is moving from convenient sample handling to scientific autonomy. Self-driving laboratories integrate robotic synthesis, automated characterization, and ML-guided decision-making within closed-loop optimization frameworks. \citet{macleod2020selfdriving} demonstrated this paradigm for thin-film photovoltaic materials, achieving relatively quick identification of optimal compositions through iterative Bayesian optimization. Their platform executed synthesis-characterization cycles faster than human researchers could analyze individual results. This not only changes the speed of discovery but also the economics of experimental exploration.

Building on the discussion in Section \ref{sec:Accelerating-high-throughput-atomic-scale-computational-screening}, we note here the work of \citet{burger2020mobile}, who extended automation to mobile robotic systems capable of operating continuously without human intervention. Over an eight-day campaign, their robotic chemist explored the photocatalyst formulation space, conducting more than 688 experiments to identify compositions with enhanced hydrogen-evolution activity. The system exhibited adaptive behavior, modifying sampling strategies based on accumulated evidence rather than following predetermined experimental plans. This adaptability is important when navigating high-dimensional parameter spaces where optimal regions cannot be anticipated from prior knowledge.

Perhaps the most ambitious integration of computational prediction with autonomous synthesis to date is the A-Lab platform. \citet{szymanski2023alab}. While the success rates are subject to discussion\cite{leeman2024challenges}, real materials were synthesized and characterized without human hands touching the equipment, as demonstrated by algorithmic predictions. 

Examples have emerged in which industrial adoption follows academic demonstrations. Contemporary Amperex Technology Limited (CATL) implemented automated sintering workstations capable of validating 200 battery material formulations daily. This reduced development cycles by eight months compared to traditional workflows. Tesla\textquotesingle s 4680 battery development leveraged automated testing systems that compressed charge-discharge characterization from 70 hours to 6 hours per sample \citep{ding2025advancing}. These implementations have demonstrated that autonomous laboratory concepts can be scaled to manufacturing-relevant levels. However, the diversity of synthesis routes and characterization techniques required across different material classes remains a limitation for wide-reaching adoption.

\subsection{Implementation and Deployment Barriers}\label{gaps-between-benchmark-performance-and-industrial-reality}

Even impressive benchmark metrics can obscure fundamental limitations for practical utility. Most ML potentials demonstrate accuracy on thermodynamic properties under equilibrium conditions. However, practical industrial applications frequently involve non-equilibrium phenomena: catalysts undergoing deactivation, membranes experiencing fouling, coatings degrading under environmental stress. Integrating time-dependent behavior into architectures designed for static property prediction remains an open challenge. Physics-informed approaches that encode kinetic pathways and mass-transfer limitations offer promising directions, although systematic validation across industrially relevant scenarios remains lacking \citep{karniadakis2021physics}.

Another issue is out-of-distribution generalization. \citet{li2025ood} systematically probed ML model performance on materials differing structurally from training distributions, finding substantial accuracy degradation for compositions involving underrepresented elements or unusual coordination environments. These insights carry significant implications for discovery workflows that explicitly target novel chemistry. If models perform well only on materials resembling those already known, the promised exploration of uncharted chemical space may prove more limited than current projections suggest.

Data heterogeneity and deficiencies in standardization compound technical challenges. FAIR (Findable, Accessible, Interoperable, Reusable) principles enjoy rhetorical endorsement throughout the materials community, but practical implementation varies across repositories \citep{draxl2019nomad}. Metadata inconsistencies, missing uncertainty quantifications, and incompatible file formats prevent cross-dataset training, which would otherwise maximize data utility. Initiatives such as NOMAD Laboratory\citep{draxl2019nomad} aim to address these limitations, though achieving community-wide harmonization requires sustained coordination that often conflicts with immediate research priorities.

Closely related to, but distinct from, data standardization is the question of computational reproducibility, which we identify in the Introduction as a precondition for verifiable and comparable AI/ML research. In materials informatics specifically, benchmark leaderboards and reported accuracies are difficult to interpret when the training/test splits, featurization pipelines, hyperparameter search, and trained model checkpoints behind a given number are not released alongside it. Two papers reporting similar headline accuracy on, e.g., a MOF property-prediction task may not be comparable at all if one used a random split and the other a scaffold-based split, or if undisclosed hyperparameter tuning budgets differed substantially. Requiring code, data splits, trained weights, and versions of software/environments as a condition of publication, as is increasingly the norm in adjacent computational chemistry venues, would materially reduce the effort currently spent reimplementing published materials-discovery workflows from incomplete descriptions.


\subsection{Transfer-Based MOF models}
\label{subsec:transfer-based-MOF-models}
Several converging developments suggest imminent transformation in how materials are conceived, synthesized, and deployed. Foundation models trained on multimodal data, encompassing crystallographic structures, spectroscopic signatures, process parameters, and scientific literature, promise unprecedented integration of experimental and computational knowledge streams \citep{ock2024unimat}. Whether such models achieve genuine scientific reasoning or merely sophisticated pattern matching remains contested, but their potential to accelerate literature synthesis and hypothesis generation is substantial.

The introduction of Transformer models, originally developed for natural language processing, has emerged as a breakthrough across diverse scientific domains \citep{vaswani2017attention}. Their self-attention mechanism enables the natural capture of both local and long-range atomic interactions that govern key properties of porous materials, particularly MOFs (see Section \ref{sec:separation-materials}). These properties include porosity, gas adsorption, and structural stability. This allows Transformers to overcome the limited receptive field and message-passing constraints inherent to GNNs. They support highly parallel training, scale efficiently to large-scale MOF datasets, and facilitate powerful self-supervised pre-training and transfer learning, thereby reducing reliance on labeled data. Moreover, they avoid rigid graph priors and flexibly adapt to diverse MOF topologies and crystal structures while providing improved interpretability through attention weights. Collectively, these advantages significantly enhance performance in property prediction, inverse design, and multi-task modeling for MOF systems. Recent advances in Transformers have shifted MOF discovery from property-specific regression models to generalized, foundation-model approaches. A key capability driving this shift is the decoupling of feature extraction from explicit 3D atomic coordinates. For instance, structure-agnostic models leveraging text-based representations (\textit{e.g.}, MOFormer) can extract robust topological and elemental features directly from MOFid strings, circumventing the computational bottleneck of 3D structure generation \citep{cao2023moformer}. Furthermore, the development of multi-modal frameworks (\textit{e.g.}, MOFTransformer) enables the simultaneous integration of localized atom-based graphs and global energy-grid embeddings. This dual-modal approach facilitates universal transfer learning across thermodynamic, kinetic, and electronic properties \citep{kang2023multimodal}. Another transformative capability is the incorporation of explicit operating conditions (\textit{e.g.}, temperature, pressure, gas type) directly into the model architecture (\textit{e.g.}, Uni-MOF) \citep{wang2024moftransformer}. This integration allows for robust cross-system generalization, such as extrapolating high-pressure adsorption behaviors strictly from low-pressure simulation data, which significantly reduces the computational overhead associated with traditional high-pressure screening.

While attention mechanisms offer interpretability that aligns well with chemical intuition by linking local chemistry to electronic properties and grid embeddings to diffusion, critical bottlenecks remain. Current models are heavily reliant on large-scale hypothetical databases (\textit{e.g.}, hMOF) that lack synthesis feasibility annotations. Consequently, algorithms frequently identify “high-performance” hypothetical candidates that are synthetically inaccessible. Additionally, while current models predict pure-component uptakes with high accuracy, their performance degrades when addressing dynamic, multicomponent industrial conditions. For example, explicitly predicting whether competitive gas separation is fundamentally driven by thermodynamic factors (especially when kinetic adsorption rates are identical) remains a complex challenge that current data-driven models struggle to capture without explicit physical constraints.

Moving forward, chemical engineers must ensure that the screening, design, and synthesis of novel materials are explicitly directed toward practical industrial deployment. At present, available datasets and machine learning models predominantly focus on predicting equilibrium gas adsorption and separation performance based on adsorption isotherms. While these thermodynamic metrics are necessary, translating a newly discovered material into a viable industrial technology requires a comprehensive evaluation of dynamic factors. Specifically, both adsorption and desorption kinetics must be integrated alongside thermodynamic data to determine the final engineering process configuration. Therefore, a critical priority is the development of end-to-end predictive models capable of bridging the gap between fundamental material properties and optimal industrial process conditions. Achieving this direct translation from material discovery to process engineering remains a significant challenge, marking a long but essential path ahead for the field. Furthermore, standardizing benchmarking protocols is essential to address prevalent issues of poor data hygiene and overfitting in materials discovery. Currently, many machine learning evaluations rely on random data splits, which often lead to data leakage due to the high structural similarity between training and test sets. To ensure fair model evaluation, researchers must adopt strict, out-of-distribution data splitting strategies, such as topology-based or scaffold-based splits, to rigorously test true generalization capabilities. Additionally, proposed deep learning architectures must be critically evaluated against well-tuned traditional baselines, rather than weak comparators, to justify their increased computational cost. Incorporating rigorous statistical tests to verify the practical significance of performance improvements is also necessary. Adopting these strict comparison protocols will ensure that models learn underlying chemical principles rather than simply memorizing dataset biases.

\subsection{Future Trajectories and Research Imperatives}\label{future-trajectories-and-research-imperatives}

Sustainability considerations must permeate materials discovery frameworks more thoroughly than current practice allows. Most optimization objectives target performance metrics, with environmental impact relegated to post-hoc assessment. Embedding lifecycle analysis directly into generative design algorithms would enable discovery of materials balancing functional requirements with carbon footprint, supply chain resilience, and end-of-life recyclability. \citet{olivetti2018,raabe2019} demonstrated that systematic consideration of these factors can identify materials with superior sustainability profiles among otherwise equivalent candidates, suggesting a significant opportunity for improvement.

The rise of LLMs as interfaces between researchers and analytical systems may prove as consequential as the underlying algorithmic advances. Natural language interaction lowers adoption barriers, enabling operators and engineers without specialized programming expertise to query complex databases, explore what-if scenarios, and interpret ML predictions within familiar conversational frameworks. Commercial offerings from multiple vendors are entering this space, though validation of LLM-generated scientific claims requires significant attention to hallucination risks, citation verification and interpretation, and overall reliability. LLMs have demonstrated substantial potential in reticular chemistry, encompassing literature mining, material design, and synthesis automation \citep{zheng_2025}. Notably, this potential extends to the targeted design and synthesis of materials tailored to specific performance requirements, enabling data-driven optimization of structure-property relationships. \citet{lai_2023} reported an innovative AI-driven workflow for catalyst optimization, with ammonia synthesis catalyst development as a case study, by integrating ChatGPT-based large LLMs with Bayesian optimization and an active learning loop \citep{lai_2023}. Specifically, LLMs were employed to mine scientific literature and construct a comprehensive chemical space for catalyst synthesis, while Bayesian optimization enabled iterative parametric optimization within this space. This integrated approach successfully predicted the optimal activation parameters for the Co\textsubscript{3}Mo\textsubscript{3}N ammonia synthesis catalyst, thereby significantly streamlining the catalyst development process and enhancing its overall efficiency.

It is critical to establish quantitative relationships between material properties and process performance metrics, including selectivity, conversion, energy consumption, and fouling rates to realize the full benefits of ML and AI materials development. This requires bridging molecular-scale predictions with process-scale models \cite{adjiman2021process}. Although progress has been made in several application areas (Sections \ref{sec:integration-of-aiml-with-multiscale-and-reactor-models} and \ref{sec:multiscale-modeling-and-design}) Multiscale integration represents the next critical frontier where materials informatics intersects with process systems engineering. This research domain would enable the co-optimized design of both materials and the processes that manufacture and deploy them, potentially unlocking efficiency gains inaccessible through sequential optimization of either domain alone. Table \ref{tab:summary-materials} summarizes the findings of this section.

\begin{longtable}{@{}p{0.30\linewidth}p{0.32\linewidth}p{0.32\linewidth}@{}}
\caption{Materials and Interfaces, and Applied Chemistry: current capabilities, remaining barriers, and future opportunities.}\label{tab:summary-materials}\\
\toprule
\textbf{Current capabilities} & \textbf{Remaining barriers} & \textbf{Future opportunities} \\
\midrule
\endfirsthead
\multicolumn{3}{l}{\textit{(continued)}}\\
\toprule
\textbf{Current capabilities} & \textbf{Remaining barriers} & \textbf{Future opportunities} \\
\midrule
\endhead
\bottomrule
\endfoot
\bottomrule
\endlastfoot
GNNs and equivariant architectures (NequIP/Allegro, MACE, M3GNet, CHGNet) predict formation energies, band gaps, and elastic moduli near DFT accuracy at a fraction of the cost & Reported benchmark accuracies (e.g., $\sim$0.2 eV MAE) are often sufficient for screening but inadequate for quantitative kinetic modeling & Physics-informed extensions of universal potentials that capture non-equilibrium/kinetic behavior beyond static equilibrium properties \\
Universal ML interatomic potentials (e.g., GNoME) combined with iterative ML--DFT cycles have expanded the known stable-materials database by an order of magnitude & Out-of-distribution generalization is weak for compositions with underrepresented elements or unusual coordination environments & Systematic, topology/scaffold-based (rather than random) benchmark splits and comparison against strong traditional baselines \\
Generative and inverse design (CVAE, transformer/foundation models such as MOFormer, MOFTransformer, Uni-MOF) propose new membranes, surfactants, and reticular materials with targeted properties & Hypothetical materials databases used for training often lack synthesis-feasibility annotations, yielding "high-performing" but unsynthesizable candidates & Coupling generative design with synthesis-feasibility screening and lifecycle/sustainability objectives embedded directly in the optimization \\
Self-driving laboratories (A-Lab, mobile robotic chemists) autonomously synthesize and characterize new materials, with industrial analogues emerging (e.g., automated battery-material sintering and testing) & Diversity of synthesis routes and characterization techniques across material classes limits broader industrial scaling of autonomous platforms & Wider industrial adoption of closed-loop autonomous discovery platforms across additional material classes \\
Standardized benchmarks (Matbench, OC20/OC22) and infrastructure (cloud HPC) support reproducible, community-wide evaluation of materials ML models & Data heterogeneity, inconsistent metadata, and incomplete FAIR-principle adoption across repositories hinder cross-dataset training & Community-wide data harmonization (e.g., via initiatives such as NOMAD) and multimodal, literature-integrated foundation models for materials discovery \\
Benchmark datasets and leaderboards report standardized accuracy metrics across many groups & Code, trained model checkpoints, exact data splits, and hyperparameter/tuning details behind reported benchmark numbers are frequently withheld, so reported accuracies cannot be independently reproduced or fairly compared across studies & Minimum reproducibility requirements (released code, splits, weights, hyperparameters, software versions) as a condition of publication, as increasingly enforced in adjacent computational chemistry journals \\
\end{longtable}

\section{Process Systems Engineering}

\subsection{Process and Product Design}
\label{sec:aiml_process_design}


Techniques such as Bayesian optimization~\cite{paulson2025bayesian,wang2022bayesian,frazier2018tutorial}, generative AI~\cite{khater2025generative,decardinelson2024generative}, and reinforcement learning~\cite{shin2019reinforcement,gu2024review} introduce new paradigms to leverage data to accelerate the exploration of complex design spaces and to generate new designs.  

Process design is part art, part science — engineers draw on knowledge and creative insight to identify designs that fulfill different functions, and generating alternatives that are feasible from economic, environmental, and safety standpoints remains a persistent challenge. A critical first step is finding suitable representations (abstractions) of the design space to be explored. Digital representations have been proposed that capture process topology, relevant phenomena, and key properties (e.g., material flows, phase equilibria, chemical reactions)~\cite{mencarelli2020review,pistikopoulos2024advanced}, enabling the construction of process databases and the development of generative AI/ML models capable of proposing new designs~\cite{vogel2023learning,schweidtmann2024mining}. As an example, \citet{jami2024ccu} proposed a LLM for carbon capture and utilization.

This is analogous to how large language models use vast text corpora to generate new narratives: data representations of process structures define a language that AI/ML can use to generate novel alternatives. These capabilities can be further combined with rigorous process simulations and active learning to identify designs with desired characteristics. Key open challenges remain, however — capturing complex logic and expert intuition, and determining whether generative AI/ML can truly extrapolate beyond its training data to achieve genuine discovery rather than recombination.

Bayesian optimization (BO) has gained significant traction as a data-driven technique for process design, capable of optimizing both topology and operating conditions~\cite{paulson2025bayesian,wang2022bayesian,frazier2018tutorial,baldea2026multiscale}. Unlike generative AI/ML, BO requires the user to more explicitly define the design space (e.g., via a superstructure of alternatives), but in return offers a principled framework for fusing experimental data with physical knowledge to steer design toward desired goals~\cite{paulson2022cobalt}. This makes BO particularly valuable for experimentalists seeking to integrate process data with techno-economic and lifecycle performance evaluation. BO has also become a dominant paradigm in experimental and product design, with successful applications in solvent selection for extraction~\cite{qin2024identifying}, electrolyte optimization for batteries~\cite{yik2025accelerating}, and product formulation~\cite{cao2021optimization}. Despite these successes, handling discrete design decisions and high-dimensional spaces remains challenging. The PSE community has made meaningful progress here, proposing discrete kernel models~\cite{thebelt2022tree,hase2021gryffin} and kernels that operate directly on graph and network structures~\cite{xie2025bogrape,paulson2022cobalt,gonzalezduque2024survey}.

In parallel, there has been growing interest in robust and safe BO for uncertainty-aware and safety-critical design. Robust BO explicitly accounts for parametric or environmental uncertainty, seeking designs that perform well under worst-case or risk-based conditions — exemplified by adversarially robust BO~\cite{paulson2022adversarially} and structure-exploiting frameworks such as BoFlex~\cite{kudva2024robust} and BONSAI~\cite{kudva2025bonsai}, which target flexibility and networked process systems under uncertainty. Safe BO, by contrast, ensures that all evaluated designs remain within a safe operating region while gradually expanding it, as in SafeOpt~\cite{berkenkamp2023bayesian} and domain-specific strategies such as safe explorative BO for plasma medicine. Together, these developments push BO beyond nominal optimization toward uncertainty-aware, safety-conscious frameworks better aligned with industrial practice.


Additional opportunities exist to leverage autoencoder models for mapping discrete decision spaces into continuous ones, enabling Bayesian optimization to conduct more effective searches. This approach has seen considerable use in molecular design, though its limitations remain poorly understood \cite{gomezbombarelli2018automatic}. A key — and sometimes underappreciated — insight from these studies is that AI/ML models inherently learn smooth, continuous latent representations from complex, high-dimensional input data. Graph convolutional neural networks, for instance, take an inherently discrete structure (a graph) and, through successive convolution operations, distill it into a continuous feature space. Transformer-based large language models similarly project discrete token representations into continuous embedding spaces. Central to this transformation is the use of probabilistic representations: variational autoencoders, for example, encode input data into a probabilistic latent space that is both continuous and structured, making it amenable to gradient-based optimization methods such as Bayesian optimization \cite{gomezbombarelli2018automatic}.


Significant open questions remain on how to learn latent representations that are easier for optimization models to navigate. The PSE community already implicitly exploits these capabilities by leveraging large language model architectures, but there is a clear need to tailor these tools to specific process and product design contexts. A particularly promising direction is using dimensionality reduction to help users visualize high-dimensional design spaces — molecular design has demonstrated this effectively, with visualizations that allow researchers to identify structural families and promising candidates~\cite{helfrecht2020structure,sorourifar2025adaptive}. Analogous tools for process design would be equally valuable: How similar or dissimilar are two designs? What families of designs exist? One could envision an interactive copilot that proposes candidates, visualizes trade-offs, and helps engineers navigate clusters of designs — moving beyond the single ``optimal'' solution. Such tools are beginning to emerge for chemical space exploration~\cite{sosnin2025chemical}, and developing analogous systems for process structure spaces is a natural and important next step.
%
%

A further challenge lies in integrating established design tools — such as superstructure optimization — with AI/ML techniques. Superstructure optimization is a powerful paradigm that encodes all possible design alternatives and leverages mixed-integer programming to navigate combinatorial spaces without repetitive simulations. However, mixed-integer approaches struggle with the complex nonlinearities arising from detailed process phenomena. The PSE community has addressed this partly through surrogate modeling, replacing specific unit operations with mixed-integer linear representations~\cite{henao2011surrogate,misener2023formulating,gupta2024datadriven}, but surrogate construction itself requires data collection via rigorous simulations. Better integration and streamlining of simulation, surrogate modeling, and design optimization workflows remains an open need.

Despite significant advances that promise to transform how design tasks are conceived and executed, substantial challenges remain in making these tools effective for practitioners. Full automation of design is unlikely — nor is it the right goal. Instead, the field should strive to build tools that help engineers pose better questions, explore alternatives more efficiently and systematically, and quantify trade-offs and uncertainty in ways that are simply not possible today.

\subsection{Process Control}
\label{sec:aiml_process_control}
Data-driven techniques have always been a key component of process control, from model discovery and controller tuning/learning, to state estimation, fault detection, and, more recently, accelerating the solution of control problems.
\subsubsection{Learning dynamic models}

Perhaps the most fundamental step in any controller design task is process modeling — constructing a model that predicts the dynamic behavior of the system. When the model structure is known, the task reduces to parameter estimation (system identification)~\cite{sung2002system, ahmed2025comparative}. When it is not, neural networks~\cite{bhat1990use, su1992long, su1993integration, cheng1995identification, hoskins1988artificial, anderson1996comparison, gonzalez1998identification, kuure2006computationally, esche2022architectures, wu2020process} or Gaussian processes~\cite{wang2025gaussian} can be used to learn approximate models. However, deep learning models lack interpretability, which limits trust in the safety-critical applications common in chemical engineering.

This has motivated sparse regression approaches~\cite{brunton2016sindy, narasingam2018data, lejarza2022datadriven}, where basis functions are used to represent the dynamics. Beyond predictive accuracy, dynamic models learned from data must also preserve key system properties — stability, dissipativity, and relevant time scales. This has driven efforts to learn control-relevant structure directly from data, including dissipativity~\cite{de2019formulas, koch2021determining, tang2019dissipativity}, Lyapunov functions~\cite{khansari2011learning, umenberger2018maximum, kolter2019learning, lawrence2020almost, shah2025lyapunov, li2026learning, mitrai2026constrained}, and Koopman operators~\cite{williams2015data, kaiser2021data}.

\subsubsection{Learning and tuning controllers}

Machine learning models have also been used to learn controllers from data. A classic example is reinforcement learning, in which a policy that maps the system's state to a control action is learned to maximize a reward \cite{hoskins1992process, spielberg2019toward, shin2019reinforcement,bloor2024control}.  If the process model is known, one can use neural networks to parameterize and design the controller using standard control-theory as well as data-driven approaches \cite{psichogios1991direct}. The dual task of learning controllers from data is that of learning observers. Observer design is a well-studied problem in process control \cite{kazantzis1998nonlinear, bernard2022observer}.

An alternative to data-driven controller design is to approximate control laws from data using supervised learning. A characteristic example is the approximation of MPC laws using neural networks \cite{kumar2018deep, hertneck2018learning,chen2022large} and decision trees \cite{drgovna2018approximate, mitrai2025discovering}, as well as learning solution operators using Deep Neural Operators \cite{nassaji2026approximate}. These approaches learn a multiparametric mapping between the system state and the optimal control action, thereby overcoming the challenges of solving nonlinear optimization problems online. Recently, several data-driven approaches have been proposed that parameterize the observer via neural networks \cite{peralez2021deep, menchetti2022linear, tang2023data, woelk2026neural}.

Finally, ML has been used to tune controller parameters. For this task, the tuning is posed as a black box optimization problem and data-driven optimization techniques, such as Bayesian Optimization, can be used to optimize the black box function, i.e., control performance \cite{lu2020mpc, sorourifar2021data}.

\subsubsection{Accelerating computation within MPC}

Beyond surrogate modeling and policy learning, a newer direction utilizes ML to accelerate the solution algorithms used to solve optimal control problems~\cite{mitrai2025accelerating}. Here, ML models learn the behavior of the numerical method itself — specifically, how to tune solver hyperparameters for a given MPC instance. Applications include accelerating Benders decomposition for mixed-integer MPC by learning warm starts~\cite{mitrai2024taking}, approximating transition costs between operating points~\cite{mitrai2024computationally}, selecting the best solution algorithm~\cite{mitrai2024taking, mitrai2025efficient}, and choosing between centralized and distributed MPC architectures~\cite{cooper2025learning}. Reinforcement learning has also been applied within Benders decomposition to navigate binary variable spaces~\cite{agyeman2025graph} and to select the level of inexactness at each iteration~\cite{li2025learning}.

These advances demonstrate that AI can accelerate MPC without compromising solution quality or stability guarantees. Notably, existing work focuses predominantly on mixed-integer MPC; extending ML-based tuning to nonlinear optimization solvers for continuous nonlinear MPC systems represents a promising open direction~\cite{chen2011random}.

\subsubsection{Learning the structure of optimal control problems}

Learning approaches have also been applied to discover the structure of control problems using network science and statistical inference~\cite{del2016automated, jogwar2017community, tang2018optimal, mitrai2022stochastic, mitrai2021efficient}. In these methods, a control or optimization problem is represented as a graph encoding interaction patterns among variables and constraints, and has been used to uncover structure in dynamical systems~\cite{jogwar2017community}, MPC problems~\cite{tang2018optimal}, and general decision-making problems~\cite{del2016automated, mitrai2022stochastic, mitrai2021efficient}. A current limitation is that these approaches consider only structural interactions; extending them to account for functional interactions among variables could yield better decompositions, particularly when solution time is the primary performance criterion.

\subsection{Supply Chain Planning, Scheduling, and Management}

Supply chain networks in the chemical, energy, and bioprocess sectors are characterized by multi-tier structures consisting of suppliers, manufacturers, distributors, retailers, and end customers, interconnected through complex logistics and information flows. Global supply-chain disruption events, such as COVID-19 pandemic-related shutdowns, silicon wafer supply disruptions following the 2011 Japan tsunami, climate risks for sustainable biomass supply, supply chain shifts caused by the 2018 U.S.\ tariffs, growing cyber-security risks in globalized networks, and rising sustainability requirements have increased the need for planning approaches that are able to capture global disruptions and their effects across supply chains \cite{katsaliaki2022supply}.

Early supply-chain models were built using transaction-focused approach that lacked coordination across functional domains, targeting cost efficiency within individual stages rather than system-wide performance \cite{min2002supply}. The best-value supply chains concept emphasizes a shift in supply-chain thinking from cost-focused, functionally segmented operations toward coordinated, system-wide performance optimization under multiple objectives, including responsiveness, flexibility, and alignment among partners \cite{ketchen2007bridging}. Much of the supply-chain modeling literature in 2000s was based on simplified modeling assumptions, including aggregated parameterizations and reduced network representations, often uses deterministic, single-stage, or single-echelon formulations to maintain tractability in large-scale supply-chain models \cite{tsiakis2001design,melo2009facility}. While these approaches were well suited for earlier or highly aggregated supply-chain settings, they offer limited flexibility for supporting the agility, adaptability, and coordination required in today's data-rich and time-varying environments.

AI/ML provide an opportunity for combining large-scale data analytics, probabilistic forecasting, and optimization-driven decision support \cite{leonard2021accelerate,venkatasubramanian2019promise}. Methods ranging from supervised learning to reinforcement learning, deep generative models, and digital-twin frameworks enable the characterization of complex causal relationships, rapid scenario evaluation, and self-optimizing operations. As organizations increasingly demand agility and resilience, ML offers the ability to react immediately to market fluctuations, changing customer preferences, transportation bottlenecks, and production disruptions. In the following subsections, we highlight several key AI/ML applications for supply chain planning, scheduling, and management.

\subsubsection{Data-Driven Forecasting for Inventory, Procurement, and Demand Management}


AI/ML methods have become widely used for demand, inventory, and procurement forecasting in modern supply chains. By capturing nonlinear relationships and complex temporal dynamics from high-dimensional data, ML techniques consistently outperform traditional statistical approaches and support more reliable demand planning~\cite{fuentescortes2022machine, kagalwala2025predictive}, contributing to lower inventory costs, reduced stockouts, and more stable production scheduling.  Beyond forecasting, statistical learning methods such as principal component analysis have been applied to monitor inventory, demand, and logistics data streams, enabling early detection of anomalies and disruption risks in multi-echelon supply chains~\cite{wang2020supply}.


A variety of ML models — including multivariate regression, neural networks~\cite{lu2012sales}, and long short term memory networks LSTMs~\cite{abbasimehr2020optimized} — have been applied to demand and inventory forecasting, consistently outperforming baseline time-series methods across industrial settings~\cite{carbonneau2008application} and improving both demand signal accuracy and operational resilience in logistics networks~\cite{thomas2021enhancing}.

%

Kernel-based methods have proven effective when upstream procurement signals influence downstream demand; in the pharmaceutical sector, for instance, incorporating supplier shipment histories and production delays into forecasting models measurably improves accuracy~\cite{zhu2021demand}. By anticipating shortages and disruptions before they propagate downstream, predictive analytics enables proactive risk mitigation and better sourcing decisions — highlighting ML's broader role in capturing cause-and-effect relationships across supply chain tiers.

Beyond forecasting, data-driven optimization frameworks based on learned uncertainty information have been developed for decision modeling. For instance, Bayesian decision-support approaches have been proposed to incorporate incomplete performance data across economic, environmental, and social criteria using Markov chain Monte Carlo simulation, enabling robust supplier ranking and scenario analysis under data scarcity \cite{sarkis2015supplier}. Ning and You developed a data-driven adaptive robust optimization approach that applies principal component analysis and kernel density estimation to capture correlations and asymmetry in uncertainty data, enabling multi-stage planning decisions that adapt to uncertainty realizations \cite{ning2018datadriven}. In this context, generative AI enables data-driven generation of demand and disruption scenarios for scenario-based analysis and decision support in supply-chain management \cite{jackson2024generative}.

Deep RL has also been applied to inventory decision-making. By using deep neural networks to learn dynamic and state-dependent decision policies, deep RL is well suited for sequential supply-chain problems under uncertainty. Deep RL can learn fully dynamic reorder policies for multi-echelon inventory systems, offering an alternative to find an optimal inventory policy under structural changes in the environment \cite{hammler2023fully}.

\subsubsection{Dynamic Scheduling, Maintenance Planning, and Operational Optimization}

In complex industrial systems, production scheduling and maintenance planning require fast adjustments in response to changing demand, equipment availability, and resource constraints. AI/ ML methods have therefore been increasingly adopted to support real-time, data-driven operational decision-making and reduce computational burden. In particular, surrogate modeling approaches have been used to approximate computationally expensive process models or mixed-integer scheduling formulations, enabling rapid evaluation of alternative production schedules and operating policies \cite{dias2019integration,luo2023multifeedstock,tsay2020integrating}. Supervised learning has been integrated with mixed-integer programming to accelerate large-scale industrial scheduling decisions, as demonstrated in crude oil scheduling for coastal refineries with long-distance pipelines \cite{qin2025crude}.

Beyond supervised learning approaches, reinforcement learning (RL) is particularly well suited to dynamic operating environments because it can learn adaptive decision policies for production sequencing, energy shifting, and inventory restocking under uncertainty. RL methods have been applied in process industries to manage competing operational targets and fluctuating resource availability \cite{bloor2025hierarchical}. Deep RL has further been explored for chemical production scheduling by formulating the scheduling problem as a sequential decision-making process under uncertainty. Hubbs et al.\ developed a deep RL framework that enables online schedule generation in response to stochastic demand and losses from production condition changes \cite{hubbs2020deep}. The method outperforms the mixed-integer linear programming (MILP) schedulers operating and having reasonable computational costs, showing the potential of real-time schedule optimization.

To facilitate the development and benchmarking of RL algorithms, the PSE community has also developed several representative benchmark environments, including OR-Gym~\cite{hubbs2020or}, SafeOR-Gym~\cite{ramanujam2025safeor}, and PC-Gym~\cite{bloor2025pc}. These platforms provide standardized environments for evaluating learning algorithms across a broad range of operations research and process systems engineering problems, thereby accelerating methodological development and reproducible performance comparisons.

From a PSE perspective, recent work highlights the importance of hybrid AI frameworks that tightly integrate data-driven learning with first-principles models and domain knowledge to enable safe, interpretable, and operationally feasible decision-making in process manufacturing systems \cite{ali2025hybrid}. For instance, Burtea and Tsay proposed a constrained reinforcement learning framework that embeds known operational constraints directly into the action-selection process through optimization over learned value functions, ensuring feasibility and safety of inventory and operational decisions \cite{burtea2024constrained}. Hybrid simulation-based optimization frameworks have also been applied to study centralized and decentralized decision-making in complex supply chains, capturing dynamic interactions and coordination effects under uncertainty \cite{singh2014biorefinery,sahay2014hybrid}.

Predictive maintenance is another key application. ML models that learn equipment degradation patterns or failure probabilities allow maintenance to be scheduled before failures occur, reducing downtime and improving asset utilization. In chemical and energy systems, Q-learning and agent-based simulation frameworks have been explored for decentralized equipment control and failure-response actions \cite{chen2025aiagent}.

\subsubsection{Transportation, Routing, and Distribution Optimization}

Transportation and warehousing activities relate to data such as vehicle movement data, shipment records, traffic information, and warehouse logs. Managing these complex and time-varying data series is crucial for improving routing efficiency, delivery reliability, and distribution planning. As a result, a variety of data-driven and learning-based methods have been developed to support decision-making in logistics networks.

Beyond traditional forecasting tasks, recent studies show that AI-based methods are increasingly used to support decision-making in transportation, logistics, and energy systems with sustainability goals. For example, AI models that combine neural networks with optimization techniques have been applied to analyze decarbonization options and operational trade-offs in industrial and maritime energy systems \cite{kazi2022practicality}. In a recent perspective article, Kamkar et al.\ emphasize the growing role of AI in sustainable resource management, supply chain planning, and logistics optimization across chemical and energy sectors \cite{kamkar2024artificial}.

RL methods have been applied to dynamic vehicle routing, last-mile delivery, and multi-agent coordination, where agents learn adaptive decision policies that respond to uncertainty and time-varying operating conditions, which is challenging to represent with fixed rules in static optimization models \cite{rolf2023review}.

When combined with digital supply chain twin frameworks, data-driven methods support the monitoring, simulation, and reconfiguration of transportation processes and logistics flows in response to disruptions, enabling improved visibility and resilience of distribution networks \cite{ivanov2021digital}. Empirical studies in retail and parcel distribution report 10--15\% transportation fuel reductions after adopting AI/ML-enhanced routing strategies \cite{vummadi2024integration}.

\subsubsection{Sustainable and Policy-Aware Supply Chain Design}

In recent years, sustainability regulations and public policies such as carbon pricing, emissions caps, trade restrictions, and reporting requirements have influenced supply-chain design and operation. These policy constraints introduce additional uncertainty, nonlinearity, and multi-objective trade-offs that are difficult to represent using static or deterministic planning models.

AI/ML methods support policy-aware supply-chain decision-making by integrating regulatory constraints and sustainability metrics directly into data-driven optimization and control frameworks. Predictive models estimate policy-relevant variables (e.g., emissions), while prescriptive formulations incorporate these estimates into sourcing, production, transportation, and inventory decisions under regulatory limits. Reinforcement learning has been explored for policy-constrained operational control, enabling adaptive responses to mechanisms such as carbon trading or emissions caps while maintaining economic performance \cite{wang2024carbon}.

Beyond numerical optimization and traditional decision-making frameworks, as discussed in the previous section, LLMs have demonstrated strong capabilities in extracting and structuring information across many domains. Thus, LLMs may provide a new pathway for policy-aware supply-chain management by enabling the automated extraction, interpretation, and organization of regulatory texts and sustainability standards from unstructured reports. A recent work demonstrates the use of LLMs with mixed-integer optimization to support explainability, scenario exploration, and decision transparency in large-scale industrial supply chains \cite{li2023llm}.

\subsubsection{Emerging AI Methods for Supply Chain Transformation}

Emerging AI methods enable transformative advances in supply-chain planning, optimization, and operational decision support. Beyond developing accurate predictive models, modern AI-enabled PSE increasingly focuses on integrating these models into closed-loop decision-making frameworks that continuously adapt to changing operating conditions. These include: (i) predictive and prescriptive analytics, where predictive models estimate future system states (e.g., demand, supplier reliability) and prescriptive formulations such as MILP, mixed-integer nonlinear programming (MINLP), or stochastic programming integrate ML-based forecasts to recommend optimal actions; (ii) digital twins, which create computational replicas of plants, logistics networks, or entire supply chains to evaluate ``what-if'' scenarios and test disruption responses, with ML accelerating validation and real-time updating \cite{escribagelonch2024digital}; (iii) generative AI methods for scenario generation which provide demand trajectories, disruption scenarios, and multi-factor risk profiles to support robust stochastic optimization \cite{francisonotole2022role}; (iv) RL for sequential decision-making, where agents learn policies for inventory management, operational control, production switching, or vehicle routing in dynamic environments; and (v) blockchain-enabled data governance, which enables decentralized, tamper-resistant data sharing across multi-stakeholder supply-chain networks. When combined with AI-driven analytics, blockchain-validated data enhance end-to-end supply-chain traceability, transparency, and strengthen trust \cite{vummadi2024integration}. Together, AI/ML methods support the development of more adaptive, resilient, and data-driven supply-chain systems.

While these AI methods provide powerful predictive and optimization capabilities, their practical value depends on successful deployment within closed-loop PSE workflows. The lifecycle of AI-enabled decision-support systems typically begins with data acquisition from process sensors, enterprise databases, laboratory measurements, and supply-chain information systems, followed by data preprocessing, feature engineering, and model identification using statistical or machine-learning approaches. Developed models should then undergo rigorous validation and uncertainty quantification to ensure prediction reliability before being integrated with optimization, scheduling, and model predictive control frameworks for operational decision making \cite{chevron2023digital} . Rather than remaining static offline models, AI systems should be continuously updated as new process and operational data become available to mitigate model drift and maintain predictive performance under changing operating conditions. Continuous monitoring of model accuracy, automated retraining strategies, and synchronization through digital twins further enable adaptive optimization and resilient real-time supply-chain management \cite{sansana2021hybrid}.

Despite rapid development, several factors may limit large-scale adoption of AI/ML in supply chains, including high data storage and processing costs, data-privacy and data-sharing constraints, organizational resistance to automation, and limited availability of high-quality data \cite{kusiak2017smart,ivanov2021digital}. Addressing these challenges requires coordinated investments in data governance, workforce development, cross-functional alignment, and scalable digital facility. Beyond technical model development, successful industrial deployment also requires AI lifecycle management, including continuous model validation, uncertainty management, monitoring of model degradation, and mechanisms for periodic model updating. At the same time, growing supply chain disruptions, geopolitical tensions, and economic uncertainties underline the need for future supply chains to evolve toward more open, reactive, and robust structure where AI models operate as continuously updated components of integrated optimization and control systems rather than standalone predictive tools. Table \ref{tab:summary-pse} summarizes the findings of this section.

\begin{longtable}{@{}p{0.30\linewidth}p{0.32\linewidth}p{0.32\linewidth}@{}}
\caption{Process Systems Engineering: current capabilities, remaining barriers, and future opportunities.}\label{tab:summary-pse}\\
\toprule
\textbf{Current capabilities} & \textbf{Remaining barriers} & \textbf{Future opportunities} \\
\midrule
\endfirsthead
\multicolumn{3}{l}{\textit{(continued)}}\\
\toprule
\textbf{Current capabilities} & \textbf{Remaining barriers} & \textbf{Future opportunities} \\
\midrule
\endhead
\bottomrule
\endfoot
\bottomrule
\endlastfoot
Bayesian optimization is widely used for process/product design, with extensions for discrete, graph-structured, and robust/safe design under uncertainty & Generative/AI process-design tools remain largely interpolative; genuine extrapolation beyond training data toward novel designs is unproven & Interactive tools/copilots that help engineers visualize, cluster, and navigate high-dimensional process-design spaces \\
ML-learned dynamic models (Koopman operators, sparse regression, neural/Gaussian-process surrogates) support system identification while increasingly preserving stability and dissipativity properties & Lack of open, shared process-design benchmark datasets (unlike molecular/materials domains) due to proprietary industrial data limits systematic algorithm comparison & Community efforts to establish open process-design benchmarks and metrics, drawing on precedents from self-driving-lab and BO communities \\
RL-based and NN-approximated MPC laws, together with ML-accelerated mixed-integer solvers (e.g., learned Benders warm-starts), speed up online control and scheduling computation & High-dimensional, discrete/mixed-integer design and scheduling spaces remain challenging for BO and RL; deep learning controllers can lack interpretability & Tighter integration of superstructure optimization with data-driven surrogate modeling; extending ML-based solver tuning to nonlinear MPC \\
AI/ML forecasting (LSTM, kernel methods) improves demand/inventory/procurement planning; RL supports dynamic scheduling, predictive maintenance, and vehicle routing under uncertainty & Ensuring safety and constraint satisfaction is difficult once RL policies operate outside their training distribution; online exploration is often unsafe in safety-critical settings & Broader deployment of safe/robust BO and constrained or offline RL for uncertainty-aware, safety-critical design and control \\
Standardized RL benchmark environments (OR-Gym, SafeOR-Gym, PC-Gym) and digital-twin/blockchain-enabled data-governance concepts support reproducible PSE algorithm evaluation and resilient supply chains & Supply-chain AI adoption is limited by data-privacy/sharing constraints, high infrastructure costs, and organizational resistance to automation & LLM-assisted extraction of regulatory/sustainability constraints into optimization formulations, and expanded digital-twin-based, closed-loop decision support \\
Benchmark RL environments provide shared, standardized testbeds for algorithm comparison & Beyond these environments, published PSE control/scheduling studies often withhold trained policy parameters, hyperparameters, random seeds, and  software versions, making reported performance hard to reproduce or compare across algorithms & Extending benchmark-environment culture into explicit minimum code-, model-, and environment-release requirements for PSE control/scheduling/design publications \\
\end{longtable}

\section{Industrial Applications of AI}


Computational intelligence in industrial operations has evolved over several decades, progressing from first-principles expert systems to contemporary data-driven AI, and fundamentally reshaping how chemical manufacturing facilities operate, optimize, and innovate. Process companies increasingly view AI as their primary lever for the next wave of transformational gains in safety, reliability, and sustainability; British Petroleum, for example, has significantly improved production and reduced unplanned shutdowns through advanced AI analytics~\cite{bp2025top}. This transformation, however, has been neither uniform nor complete, with some operational domains near-fully integrated while others remain largely untouched. In this section, we examine AI adoption across different operational levels and roles in process industry settings, exploring the heterogeneous adoption landscape, barriers to broader deployment, and the future trajectory of intelligent process systems.

\subsection{Historical Evolution and Current State of AI/ML in Industrial Operations}
%

\begin{figure}[htbp]
  \centering
  \includegraphics[width=0.8\textwidth]{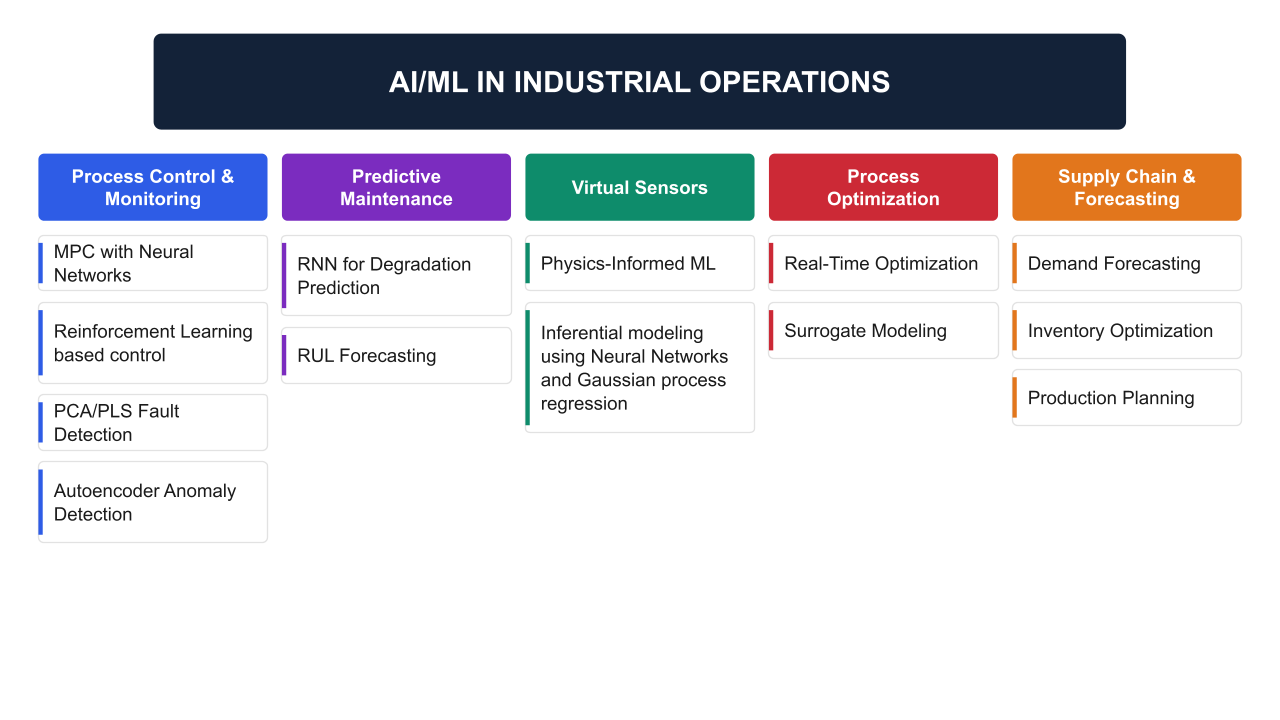}
  \caption{Application areas of AI/ML in industrial operations with substantial AI penetration.}
  \label{fig:ai_applications}
\end{figure}

Historically, high-fidelity first-principles models, such as CFD, finite element analysis, and kinetic modeling have been the workhorse of computational analytics in the process industry, enabling accurate simulation of complex ``what-if'' scenarios  (Figure \ref{fig:ai_applications}). However, their development remains time-consuming and requires specialized skills. This has spurred the adoption of AI/ML-based tools across operational domains, especially as ML algorithms, computing power, data infrastructure, and sensing capabilities have matured. The past several years have witnessed a meaningful ``democratization'' of AI/ML analytics: cloud infrastructure has made substantial computational resources accessible without major capital investment, open-source frameworks have lowered barriers to model development, and self-service platforms now enable process engineers with limited data science training to build, validate, and deploy models. This democratization has accelerated AI/ML adoption across multiple operational domains, though penetration depth remains highly uneven. Figure \ref{fig:ai_applications} shows the AI ML applications with the most penetration.

\subsubsection{Current Mainstream Applications of AI/ML in Industrial Operations}

Process control and monitoring represent the most mature application domain for AI/ML technologies in chemical plants \cite{raven2024machine}. Historically, advanced process control (APC) systems have utilized  techniques like FIR (finite-impulse response) and state-space modeling to build the model predictive controller (MPC). A recent trend is to move beyond traditional MPC implementations to incorporate advanced ML for better handling of process nonlinearities, adaptation to feedstock variations, and maintaining optimal operation across wide operating envelopes. These systems leverage techniques such as neural networks for nonlinear model identification \cite{aspentech2025aspen,balakrishnan2024neural,hammond2025selective} and reinforcement learning for adaptive control policy optimization \cite{bloor2024control,yokogawa2022press}. Complementing advanced control, AI-based tools have become increasingly prevalent in process monitoring applications. Classical techniques such as principal component analysis (PCA) and partial least squares (PLS) provide dimension reduction and fault detection in high-dimensional sensor spaces \cite{kumar2020datadriven}. Modern autoencoder neural networks learn compressed representations of normal operating conditions, with reconstruction errors serving as anomaly indicators \cite{kanno2022deep}. These monitoring systems serve as an early warning capability, detecting subtle deviations from normal operation before they escalate into process upsets, safety incidents, or product quality excursions.

Predictive maintenance (PdM) has emerged as another domain where AI/ML has achieved substantial industrial traction. The fundamental value proposition, transitioning from time-based preventive maintenance to condition-based strategies, has proven compelling across asset-intensive industries \cite{tsallis2025applicationwise}. ML models can absorb and integrate diverse, heterogeneous data streams including vibration signatures, thermal imaging, acoustic emissions, lubricant analysis, and process operating history to predict anomalous behavior \cite{jain2022creating}, equipment degradation and forecast remaining useful life \cite{liu2025comprehensive}. Techniques ranging from random forests and gradient boosting for failure classification, to recurrent neural networks for time-series prediction of degradation trajectories have demonstrated significant economic value through reduced unplanned downtime, optimized maintenance resource allocation, and extended asset lifecycles \cite{yedla2025advancing}. Companies like Shell \cite{insightsglobal2025ai} and Marathon Petroleum \cite{marathon2025ai} have reported substantial reduction in unplanned outages using ML-based PdM solutions.

Virtual (or soft) sensors are an impactful application of ML, addressing the persistent challenge of obtaining physical measurements of critical process variables. Virtual sensors are implemented using neural networks, Gaussian process regression, or ensemble methods and are used to infer hard-to-measure variables such as product composition, reaction conversion, or catalyst activity from readily available temperature, pressure, and flow measurements \cite{kadlec2009datadriven}. Recent advances in physics-informed ML modeling, which embed known physical constraints and conservation laws into ML algorithms, have significantly improved the extrapolation capabilities and physical consistency of these models \cite{velioglu2025physics}.

Continuous process optimization is another important domain which is witnessing increasing AI adoption \cite{su2025machine}. Rather than periodic offline optimization studies followed by manual implementation of recommended operating conditions, modern real-time optimization (RTO) systems perform closed-loop optimization where machine learning models serve as surrogate representations of complex process economics and constraints \cite{raven2024machine}. These surrogates enable rapid evaluation of alternative operating strategies and real-time optimization in response to changing feedstock properties, energy prices, and product demands.




\subsubsection{Persona-based Applications}

A critical insight that has emerged from both successful and failed AI/ML deployments in process industries concerns the necessity of persona-centric design. The recognition that plant operators, process engineers, maintenance personnel, and plant managers operate with different decision-making contexts, has profound implications for how AI/ML systems should be constructed and how their outputs should be presented. The governing principle of ``delivering the right information, at the right time, to the right people'' sounds deceptively simple but proves challenging to implement in practice.

Plant operators, who manage real-time process behavior and respond to disturbances and upsets, require AI/ML systems that provide immediate, actionable intelligence. Anomaly detection alerts must indicate not merely that a deviation has occurred, but what specific process variables are implicated and what operational interventions might address the issue. The critical requirement for operator-facing AI systems is transparency and explainability; operators must understand why a recommendation is being made and have confidence in the underlying reasoning.

Process engineers engage with AI/ML systems at a different level of abstraction. Their responsibilities encompass performance optimization, troubleshooting recurring issues, and continuous improvement of process operations. AI tools for this persona must support more sophisticated analyses: automated root cause analysis that systematically evaluates potential contributors to process upsets, full plant modeling capabilities that enable quantitative evaluation of proposed process modifications, and knowledge retrieval systems that surface relevant historical learnings from previous optimization studies or equipment failures. The temporal scale is longer (hours to weeks rather than seconds to minutes) and the emphasis shifts from reactive response to proactive improvement.

Plant managers require yet another perspective on plant operations. Their decisions concern resource allocation, production planning, and risk management. AI/ML systems serving this persona must aggregate operational data into strategic insights: performance dashboards that track key performance indicators against targets and historical trends, predictive analytics that forecast production capabilities and identify bottlenecks, and risk assessment frameworks that quantify exposure to various operational and market risks. The challenge lies in distilling the complexity of plant operations into concise, interpretable representations that support executive decision-making without overwhelming users with operational minutiae.

This persona-based architecture implies that a single monolithic AI system is unlikely to serve all stakeholders effectively. Rather, a layered information architecture is required where foundational AI capabilities related to predictive models, optimization engines, and knowledge bases are accessed through persona-specific interfaces and workflows that present information in forms appropriate to each user's role, responsibilities and decision context.

\subsubsection{Recent Trends}


\textbf{Real-time digital twins:} While precise definitions tend to vary based on context, digital twins can be broadly viewed as virtual replicas \cite{chevron2023digital} of any equipment, process, or entire plant. Real-time digital twins that can run in parallel to the physical assets provide unprecedented visibility into system-wide health status and efficiency. Moreover, high-fidelity digital twins that can be executed in real-time allow operators to evaluate potential control actions through what-if scenario analysis before implementation. However, owing to the challenges associated with developing a high-fidelity dynamic model whose behavior matches that of a physical plant, and which can be updated in real-time robustly with latest plant data, deploying comprehensive digital twins for complex processes still remain an aspirational target for many industries.

Nonetheless, the number of publicly reported successful digital twin applications is increasing; for example, Chevron \cite{chevron2023digital} described how digital replicas of their plants are helping  operators assess plant health in real-time. These digital twins invariably are hybrid in nature, wherein mechanism-driven first-principles equations are integrated with AI/ML techniques, resulting in computationally tractable models that are grounded in the physical realities of the underlying process.

\textbf{Hybrid Modeling:}  The broader hybrid modeling technique underpinning these industrial digital twins spans many plant-scale case studies that illustrate its maturity. Sansana et al. surveyed hybrid modeling practice across the process industries in the context of Industry 4.0, cataloging architectures, tooling, and the challenges of embedding hybrid models within digital infrastructure at scale \cite{sansana2021hybrid}; the same group later showed that hybrid architectures can improve extrapolation and transfer learning when a process shifts to a new operating regime with only limited new data, using simulated biodiesel production as a case study \cite{sansana2024extrapolation}. Shah et al. combined a deep neural network with a fermentation kinetic model to identify time-varying dependencies among kinetic parameters in an industry-scale (more than 100,000-gallon) fermenter, improving both accuracy and robustness relative to the mechanistic model alone \cite{shah2022fermentation} as an early example of the kind of plant-scale hybrid deployment envisioned above. Two recent reviews take a broader methodological view: Sharma and Liu classify hybrid approaches by whether ML complements first-principles science or science complements ML, spanning serial, parallel, and inverse hybrid architectures \cite{sharma2022hybrid}, while Schweidtmann et al. propose a more precise definition of hybrid model structures and revisit training methodologies in light of recent advances in physics-informed machine learning \cite{schweidtmann2024review}. A step-by-step tutorial by Shah et al. addresses the practical gap between these conceptual reviews and implementation, walking through the mathematical formulation and code-level construction of hybrid models for new practitioners \cite{shah2025tutorial}. Early comparative work by Feyo de Azevedo et al. anticipated much of this industrial activity, established that hybrid mechanistic–neural-network structures can outperform either a purely mechanistic or a purely data-driven model on validation data using fed-batch fermentation as a benchmark case \cite{feyodeazevedo1997hybrid}. Finally, at the kinetics scale underlying many of the reactor and catalytic applications discussed in Section 3, Pahari et al. combined machine learning with first-principles kinetic Monte Carlo simulation to define and predict “effective” activation energies on electrocatalyst surfaces that capture many-body adsorbate interactions absent from conventional kinetic models \cite{pahari2025kinetic}, illustrating that the same hybrid modeling paradigm extends from plant-wide digital twins down to the elementary reaction step.

\textbf{Prescriptive analytics} leverage AI/ML to not only predict future states but also recommend optimal actions for process improvement \cite{lepenioti2020machine}. Unlike descriptive and predictive analytics, which forecasts outcomes or flags potential equipment/process failures, prescriptive analytics recommends in real time the specific process interventions / maintenance actions that need to be taken to ensure safe and the most efficient operation in a given scenario. Specific solutions may involve usage of digital twins or ML algorithms that mine historical data to extract relevant process patterns to find optimal operational strategies.  


\subsection{Technical and Organizational Challenges to Widespread Adoption}

\begin{figure}[htbp]
  \centering
  \includegraphics[width=0.8\textwidth]{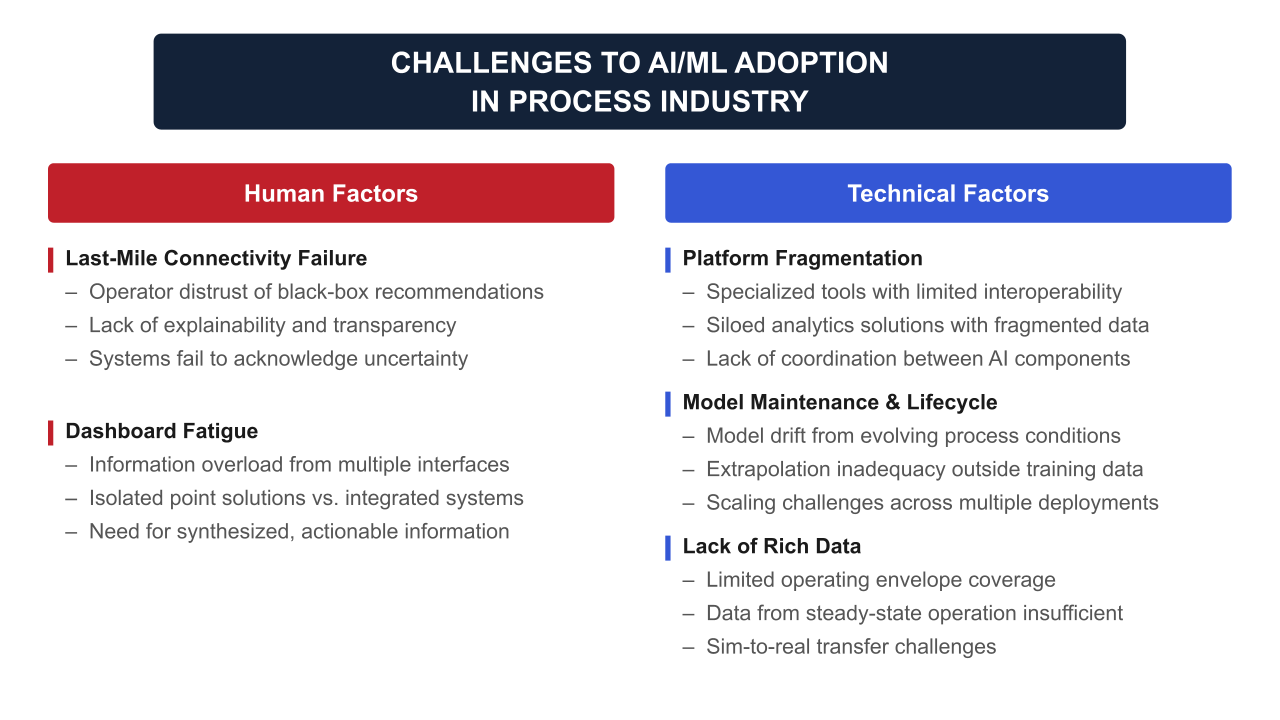}
  \caption{Factors inhibiting widespread adoption of AI/ML solutions.}
  \label{fig:ai_challenges}
\end{figure}

Despite the demonstrated value of AI/ML applications in specific domains, widespread adoption across the process industries faces substantial challenges (Figure \ref{fig:ai_challenges}). Many AI/ML applications are still treated as ``add-ons'' rather than essential components of the process operation toolkit. These challenges encompass both technical and human factors that prove equally consequential for AI/ML adoption.

The ``last-mile connectivity failure'' phenomenon, where sophisticated AI systems are developed and deployed but fail to influence operational decisions, represents perhaps the most significant barrier to realizing value from AI/ML investments. Plant operators, the ultimate end-users of many AI systems, often express distrust of black-box recommendations, particularly from systems whose internal reasoning is opaque. Building trust requires AI systems that can explain their recommendations in terms operators understand, that demonstrate consistent reliability over extended periods, and that gracefully handle situations outside their training distribution by acknowledging uncertainty rather than offering spuriously confident/hallucinatory predictions. 

``Dashboard fatigue'' has emerged as a related challenge. Plant control rooms already feature numerous displays, alarm systems, and monitoring interfaces. The introduction of additional AI-powered dashboards, regardless of their analytical sophistication, often meets resistance from operators who view them as adding to information overload rather than reducing it. A commonly heard complaint is having too many dashboards to monitor, that reflects a deeper issue: the proliferation of isolated point solutions rather than integrated systems where different AI components communicate and coordinate. What operators need is not more information but better information, synthesized and presented in forms that directly support their decision-making rather than requiring them to integrate across multiple disconnected systems.

Platform fragmentation presents another systemic challenge. The current landscape features numerous specialized AI/ML tools and platforms, each addressing specific applications, such as optimization, digital twin, production scheduling, control, monitoring, predictive maintenance, quality prediction, but with limited interoperability. This fragmentation creates silos where insights generated by one system cannot inform decisions in another. A truly intelligent plant requires integration across these boundaries such that different AI components can communicate and coordinate; for example, predictive maintenance systems can inform optimization algorithms about impending equipment constraints, and anomaly detection systems can trigger automated root cause analysis workflows.

Model maintenance presents particularly vexing challenges that compound as AI/ML deployments scale across multiple applications and plant units. Machine learning models learn relationships from historical data reflecting specific operating conditions and equipment states, but when process behavior evolves, whether gradually through catalyst deactivation and heat exchanger fouling, or abruptly through equipment modifications or feedstock changes, model predictions drift from reality \citep{hammond2025selective}. The extrapolation inadequacy inherent in most machine learning methods becomes especially problematic when processes operate outside the historical training distribution, leading to inaccurate predictions or false alarms even when process behavior remains normal for new conditions. Determining when model updates are required, distinguishing genuine model degradation from transient disturbances, and ensuring retraining improves rather than degrades performance requires sophisticated diagnostics and often human judgment. Managing these tradeoffs at scale, across dozens or hundreds of models deployed throughout a plant or across multiple facilities, becomes daunting.

The model life cycle management in practice benefits from an explicit, repeatable framework rather than ad hoc intervention. Such a framework typically comprises: (i) continuous model monitoring, tracking input/output distributions, prediction residuals, and performance metrics against a validation baseline; (ii) automated drift detection, using statistical distance metrics (e.g., population stability index, Kolmogorov–Smirnov statistics, or reconstruction error from an autoencoder trained on the reference distribution) to flag meaningful departure from the training domain; (iii) defined retraining triggers, specifying quantitative thresholds (sustained residual growth, drift-metric exceedance) at which retraining is initiated, rather than a fixed calendar schedule that may under- or over-correct; (iv) mandatory revalidation after known plant changes such as catalyst replacement, equipment modification, feedstock change must be treated as checkpoints independent of whether automated drift metrics have fired; (v) uncertainty-aware recommendations, so outputs carry calibrated confidence estimates and downstream decisions can be gated on that uncertainty; and (vi) governance that assigns clear ownership and approval authority for model updates in safety-critical applications, mirroring change-management practice already used for other safety-instrumented systems. Embedding these elements in the unified analytics platforms discussed above would let drift detected in one model (e.g., a soft sensor) automatically inform the retraining priority of downstream models (e.g., an RTO surrogate) that depend on it, rather than requiring independent, manually triggered maintenance across dozens of disconnected deployments.


The challenge of inadequate training data richness fundamentally limits the scope of achievable AI/ML applications, particularly for optimization and control applications requiring understanding of process behavior across broad operating envelopes. While chemical plants generate enormous data volumes, the vast majority accumulate during normal closed-loop operation where controllers maintain variables near setpoints, providing data valuable for monitoring applications but inadequate for optimization solutions requiring extensive coverage of the operating space. Models trained on such limited-excitation data may interpolate reasonably within narrow historical operating envelopes but fail when optimization algorithms suggest alternative strategies. The alternative of augmenting plant data with high-fidelity simulation-generated training data introduces substantial challenges: developing and maintaining comprehensive simulation models requires substantial engineering effort (often even more than training data-driven models) and ensuring models trained on simulated data perform reliably on actual plants remains nontrivial due to plant-model mismatch and unmeasured disturbances.


\subsection{Large Language Models: A Paradigm Shift in Human-AI Interaction in Process Industry}

Among the recent computational innovations, the one with potential for the most (constructive) disruption in industrial operations is the application of LLMs as ``co-pilots'' or ``virtual assistants'' to supplement the capabilities of plant operators and engineers. The claim is not that the LLMs will make existing AI applications obsolete, rather LLMs can help bring these AI applications closer to the end-users who are not proficient in advanced data analytics (as is the usual case in plant control rooms). Several commercial offerings are already appearing in this space \cite{appliedcomputing2025orbital,honeywell2025assistant}.  Below are a few ways in which LLMs can help in the operational domain.


Generative AI can enable operators and engineers to interact with complex analytical systems through natural language, dramatically lowering the barrier to adoption. Rather than learning specialized software interfaces, users can ask questions, request analyses, and explore what-if scenarios using familiar conversational interactions. For example, a plant operator, upon detecting an unusual drift in a temperature signal, could ask in plain English, ``which other signals are showing similar drifts as the reactor outlet temperature?'' and the LLM can interface with an analytics tool to provide a direct answer supported with necessary visualizations. The operator could continue the conversation to further investigate and establish if the temperature drift is an isolated phenomenon indicating a potential sensor issue or a more serious process upset. Assuming that this interface can correctly reflect physical reality, the positive impact on operator productivity could be quite high. 

The capability for contextual knowledge integration represents an equally significant advance. Large language models can in principle be fine-tuned or augmented through retrieval mechanisms to access plant-specific documentation, operating procedures, troubleshooting guidance, operator logbook entries, maintenance work-orders, equipment manuals, and historical incident reports. When combined with real-time process data and AI-generated insights, these systems can serve as intelligent assistants that provide contextualized guidance tailored to specific situations.

Conversational AI interfaces offer a potential solution to the aforementioned dashboard fatigue. Rather than presenting users with static displays that present information regardless of immediate relevance (a ``push'' model), conversational systems enable a ``pull'' mode of operation, where users request specific information as needed, in the form most useful for their current decision context. This shift from monitoring multiple parallel information streams to focused, query-driven information may prove more effective in supporting operational decision-making. The system becomes a collaborative partner that responds to expressed needs rather than a broadcasting station that demands continuous attention.

Knowledge transfer and retention of expert domain knowledge represent challenges that take on increasing urgency as experienced operators and engineers approach retirement. The accumulated wisdom (subtle indicators of equipment degradation, contextual factors affecting process behavior, troubleshooting heuristics developed over decades) resides largely in the tacit knowledge of personnel rather than in documented procedures. As opined by Baldea et al.~\cite{baldea2025from},
LLMs could utilize transcriptions of human conversations (e.g., among operators in an online meeting) as context for better decision making; the objective would be to be able to capture operational wisdom inherent in these plant control room discussions. Overall, this would closely emulate the way humans learn from each other.
It is useful to distinguish two roles LLMs can play here, since they carry very different reliability requirements. As interfaces and orchestrators, LLMs translate natural-language requests into calls to existing, verified tools such as process historians, simulators, optimization solvers, validated analytics models and translate the results back into natural language; correctness remains the responsibility of the underlying tool, and the LLM's role is interpretation and routing. As scientific or engineering reasoning engines, by contrast, LLMs are asked to perform the calculation, diagnosis, or design step themselves, drawing on patterns learned during pretraining rather than a verified computational tool. This second role is considerably riskier in process systems as LLMs carry no intrinsic guarantee of numerical accuracy, mass/energy-balance closure, or physical consistency, and can produce fluent, confident, and superficially plausible answers that are quantitatively wrong or cite nonexistent sources. LLM and agentic AI systems should, wherever possible, be restricted to the orchestrator role in operational settings connected to verified simulators, historians, and optimization tools, with a human-in-the-loop check before any recommendation is acted upon in a safety-critical decision. Recent multi-agent frameworks for process systems engineering illustrate this pattern, coupling LLM-based orchestration with specialized, verifiable computational agents rather than asking a single language model to reason end-to-end about plant behavior \cite{lima2026improving, rupprecht2026multi}.

Important caveats accompany these promising capabilities. Current LLMs can generate plausible-sounding but factually incorrect responses (so-called hallucinations) which pose obvious risks in safety-critical process operations. The tendency of these models to exhibit a confident tone even when their knowledge base and information have high uncertainty, makes detecting unreliable outputs challenging.

\subsection{Future Trajectory: Towards Increasingly Autonomous Operations}

The evolution of AI/ML applications in process industries points toward several interconnected developments that collectively promise to move operations toward more autonomous, adaptive, and intelligent operations with synergistic human-machine collaboration. As previously discussed, LLM-based solutions are poised to spearhead human-machine collaboration. As far as autonomous operations (which entail utilization of new enabling technologies to automate some practically important tasks that have hitherto relied extensively on human intervention \cite{baldea2025from}) are concerned, it is not far-fetched to speculate that in the near-to-medium term, the following technologies will become mainstream and help unleash the next wave of productivity gains.


\textbf{Unified Data Analytics Platform}

We previously discussed how operations personnel are flooded with siloed analytics solutions with fragmented data sources. Different areas of interest such as plant health and reliability, control, strategy, sustainability, optimization, albeit all correlated, have been approached individually and have their own solutions, often utilizing their own custom analytics infrastructure and rarely incorporating priorities beyond their own. There is a rejuvenated interest in the industry to move away from these custom, fragmented solutions toward unified platforms. A unified enterprise-wide platform and AI will allow a standardized way to access data across different solutions enabling a truly ``intelligent'' plant whose different components can communicate and coordinate. Achieving this level of integration requires not only technical standards for data exchange but also architectural frameworks that support loosely coupled systems with well-defined interfaces.

\textbf{Reinforcement Learning (RL)-Based Adaptive Control}

At the outset, the task of process control appears to be a mature task with established automation tools doing the bulk of the job (using classical PID controllers, MPCs, etc.). However, plant engineers will vouch for the daily challenges (due to tuning issues, creeping plant-model mismatch, etc.) they face in maintaining these systems and ensuring optimal controller performance over time. The successes of reinforcement learning in computational science augur well as a potential solution to this challenge wherein an automated RL agent continuously makes small interventions to the process to find optimal rewards and ``learn'' about the process environment on the fly to improve its actions. For example, Blum et al.\ investigated an RL controller at a Linde air separation plant and demonstrated a stable control behavior with superior control quality for over 20 days \cite{blum2021investigation}.   Even with these advances, deploying reinforcement learning directly onto physical chemical plants remains difficult. This reality requires a more cautious perspective than simply assuming RL's successes in other fields will easily translate here. Online exploration is frequently unsafe in safety-critical settings, historical data often lack sufficient variation (Section 8.2), and it is hard to guarantee closed-loop stability or constraint satisfaction once a policy moves beyond its training data. These algorithms typically require thousands to millions of environment interactions to converge, which is computationally expensive. Researchers are pursuing three complementary strategies to address these hurdles. Safe RL methods limit the exploration or action space itself through chance-constrained formulations or action-projection layers, ensuring that only feasible actions are ever applied to the plant  \cite{shin2019reinforcement, bloor2024control}. Offline RL avoids the risks of online exploration altogether by training policies entirely on historical or simulated data; recent studies have shown this works for chemical reactor control by pairing an offline policy with a real-time safety layer that corrects actions using a learned cost model \cite{durkin2025offline}.  Embedding prior knowledge of system dynamics and hard constraints directly into the learning loop is a promising route to reduce this burden. For example, RL-assisted MPC architectures constrain the RL agent's search to actions that a model-based controller has already deemed feasible, \cite{zanon2020safe} This hybrid setup drastically lowers the sample complexity compared to model-free RL baselines and enables faster convergence during training while maintaining robust constraint satisfaction. RL-based adaptive PID controller tuning solutions have also been proposed \cite{badgwell2021uspatent}.  Across all these strategies, the most persistent challenge is bridging the simulation-to-plant gap. Policies developed on simplified or first-principles simulators often underperform on real equipment because of unmodeled dynamics and plant-model mismatch. For now, RL is not a replacement for MPC but a very valuable addition to MPC and other established control strategies in safety-critical chemical processes.

\textbf{Digital Twin Integration}

Comprehensive digital twins that combine first-principles models, data-driven surrogates, and real-time optimization will enable true predictive operations and not just anticipating failures, but proactively adjusting operations to prevent issues before they manifest. Real-time digital twins running in parallel to their physical counterparts and ``keeping'' a check on the behavior of the actual operation would help avoid ``surprise'' plant upsets leading to costly plant trips. With the digital twins as components of the unified analytics platform, certain well-characterized operational regimes may become fully autonomous, with AI systems managing startups, shutdowns, and transitions while human operators focus on higher-level supervision and handling of off-normal situations.

\textbf{Holistic Closed-Loop Plant-Wide Optimization}

Process optimization is a computationally expensive task, with large complexes often broken down into smaller units for optimization purposes. Or, in a common industrial setting, MPC keeps operations at optimal control points, with optimal setpoints coming down from the RTO layer. Often, the models used by MPC and RTO layers are different, resulting in potentially sub-optimal operations especially during fast-paced process transitions. However, ongoing computational innovations \cite{du2015timescale}, development of ML-enabled digital twins, and setup of unified analytics platforms raise hope for a future where large complexes can be holistically optimized as an integrated system balancing process mode variations, product demand fluctuations, and aging equipment. Table \ref{tab:summary-industrial} summarizes the findings of this section.

\begin{longtable}{@{}p{0.30\linewidth}p{0.32\linewidth}p{0.32\linewidth}@{}}
\caption{Industrial Applications of AI: current capabilities, remaining barriers, and future opportunities.}\label{tab:summary-industrial}\\
\toprule
\textbf{Current capabilities} & \textbf{Remaining barriers} & \textbf{Future opportunities} \\
\midrule
\endfirsthead
\multicolumn{3}{l}{\textit{(continued)}}\\
\toprule
\textbf{Current capabilities} & \textbf{Remaining barriers} & \textbf{Future opportunities} \\
\midrule
\endhead
\bottomrule
\endfoot
\bottomrule
\endlastfoot
Process control and monitoring (ML-augmented APC, PCA/PLS, autoencoder-based anomaly detection) is the most mature AI/ML application domain in chemical plants & Many AI/ML applications remain ``add-ons'' rather than essential components of the operating toolkit, limiting sustained integration & Embedding AI capabilities as core, trusted components of standard operating and control infrastructure \\
Predictive maintenance using vibration, thermal, acoustic, and process-history data has delivered documented reductions in unplanned downtime at companies such as Shell and Marathon Petroleum & ``Last-mile'' adoption failure: sophisticated models often fail to influence operator decisions due to distrust of opaque, black-box recommendations & Explainable, uncertainty-aware AI systems that build operator trust and gracefully acknowledge out-of-distribution situations \\
Virtual/soft sensors (NN, Gaussian process, ensemble methods), increasingly physics-informed, infer hard-to-measure variables from routine measurements; ML surrogates enable real-time optimization (RTO) & Model drift from catalyst deactivation, fouling, and equipment changes requires ongoing maintenance; ``dashboard fatigue'' and platform fragmentation hinder adoption at scale & Formalized model-lifecycle-management frameworks (continuous monitoring, automated drift detection, defined retraining triggers, governance) embedded in unified analytics platforms \\
Hybrid digital twins (mechanistic + ML) are deployed at plant scale for real-time health monitoring  and have a growing base of plant-scale case studies across process industries & Training data reflect narrow, closed-loop operating envelopes, limiting applicability to optimization/control tasks requiring broader excitation; comprehensive high-fidelity digital twins remain aspirational for many processes & Expanded hybrid digital twins enabling autonomous management of startups, shutdowns, and transitions, with humans focused on supervisory oversight \\
LLM-based copilots are beginning to lower the barrier to natural-language interaction with plant analytics, contextual knowledge retrieval, and operator knowledge transfer & LLM hallucination, confident-but-incorrect outputs, and lack of intrinsic numerical/physical guarantees pose safety risks if used beyond an orchestrator role & Restricting LLMs to orchestrator roles interfacing with verified simulators/historians/optimizers, with human-in-the-loop review before safety-critical actions \\
Published industrial case studies (e.g., digital twins, predictive maintenance, RL-based control) report quantified operational benefits & Proprietary data and IP concerns mean industrial AI/ML studies rarely release code, trained models, or full data-processing pipelines, making reported gains difficult for independent researchers to verify or reproduce & Development of anonymized benchmark problems and reporting norms that let industrial studies disclose methodology and reproducibility artifacts without exposing proprietary plant data \\
\end{longtable}

\section{Conclusions}


Artificial intelligence (AI) and machine learning (ML) have transitioned from niche analytical tools to foundational components of modern chemical engineering research and practice. This review has surveyed recent advances across length and time scales, from atomistic simulations and materials discovery to process systems engineering and industrial operations, highlighting both transformative successes and persistent challenges. Collectively, the contributions reviewed here demonstrate that the impact of AI/ML in chemical engineering is no longer speculative: it is measurable, growing, and increasingly indispensable.

A central theme emerging across all application domains is the shift from purely data-driven models toward \emph{hybrid and physics-informed frameworks}. As emphasized in the Introduction, chemical engineering problems are constrained by mass, energy, momentum, and thermodynamic laws, and successful AI/ML models increasingly reflect this reality. Physics-informed neural networks, hybrid mechanistic--data-driven models, thermodynamically consistent learning architectures, and structure-preserving control models illustrate how embedding physical structure improves predictive accuracy, interpretability, extrapolation behavior, and safety. These developments directly address long-standing concerns about black-box learning in safety-critical and extrapolative regimes.

At the molecular and materials scales, machine-learned interatomic potentials, graph neural networks, and generative models have dramatically expanded the accessible design space for catalysts, adsorbents, membranes, and functional materials. By enabling close to \emph{ab initio} accuracy at orders-of-magnitude lower computational cost, these models have accelerated materials screening, enabled inverse design, and facilitated autonomous discovery platforms that combine computation and experimentation. Symmetry-aware and equivariant architectures have improved data efficiency and physical fidelity, while benchmark datasets and standardized evaluation protocols have strengthened reproducibility and promoted community-wide, collaborative progress.

At the process unit and plant scales, AI/ML methods have enhanced multiscale simulation, optimization, and control. Learning-based surrogates and structure-discovery methods now complement traditional first-principles approaches, enabling real-time optimization, adaptive experimentation, and closed-loop operation. In process systems engineering, Bayesian optimization, reinforcement learning, and generative models offer new paradigms for design, scheduling, and supply-chain management under uncertainty. When integrated with first-principles constraints, these approaches retain feasibility and interpretability while enabling rapid exploration of complex decision spaces.

Industrial deployment highlights both the promise and the limits of current AI/ML technologies. Applications in process monitoring, predictive maintenance, soft sensing, and optimization have delivered tangible value, yet challenges remain in last-mile adoption, model lifecycle management, trust, and organizational integration. The emergence of large language models (LLMs) as interfaces and copilots represents a qualitative shift in human--AI interaction, with the potential to reduce cognitive load, mitigate dashboard fatigue, and democratize access to advanced analytics. However, ensuring factual grounding, uncertainty awareness, and safe operation remains essential for their use in industrial environments.

Looking forward, several research imperatives stand out. These include: (i) deeper integration of conservation laws, thermodynamics, and stability guarantees into learning architectures; (ii) robust uncertainty quantification and out-of-distribution detection for safe decision-making; (iii) scalable multiscale frameworks that bridge materials discovery with process-level performance; (iv) unified, interoperable platforms that support coordinated analytics across the enterprise; and (v) explicit, enforced standards for computational reproducibility, including public release of code, preprocessing and featurization scripts, data splits, trained model parameters, hyperparameters and tuning procedures, random seeds, and software/environment versions. On this last point, we echo the emerging norms of adjacent computational disciplines and journals and argue that chemical engineering AI/ML research, and the journals that publish it, should move toward comparable minimum requirements: the current inconsistency in code and data availability has repeatedly forced independent researchers to spend substantial effort reimplementing or guessing at published workflows, and it undermines fair comparison of reported performance across studies. Progress along these dimensions will determine whether AI/ML evolves from a powerful set of tools into a truly enabling infrastructure for chemical engineering, and will be further supported by (vi) expanding availability of open-source data sets.

We conclude that AI and ML are not replacing the core principles of chemical engineering; rather, they are amplifying them. The most successful approaches respect physical laws, leverage domain knowledge, and place human expertise at the center of the decision-making loop by providing the correct level of information for the context. As the field advances toward increasingly autonomous, adaptive, and sustainable systems, the thoughtful integration of AI/ML with first-principles understanding will be essential to realizing their full potential across research and industrial practice.

\section*{Acknowledgments}
M.B. thanks Joshua Hammond, Jihun Jeong and Jong Hyun Rho of The University of Texas at Austin for their help in converting the documents submitted by multiple authors to \LaTeX, and generating and checking the bibliographic reference database.
Generative AI tools, including Copilot (\url{https://copilot.microsoft.com}) and Claude (\url{https://www.claude.ai}) were used to harmonize the text written by multiple authors. The output was verified by the authors and confirmed to be accurate.


\clearpage

\begin{figure*}
\centering
\includegraphics{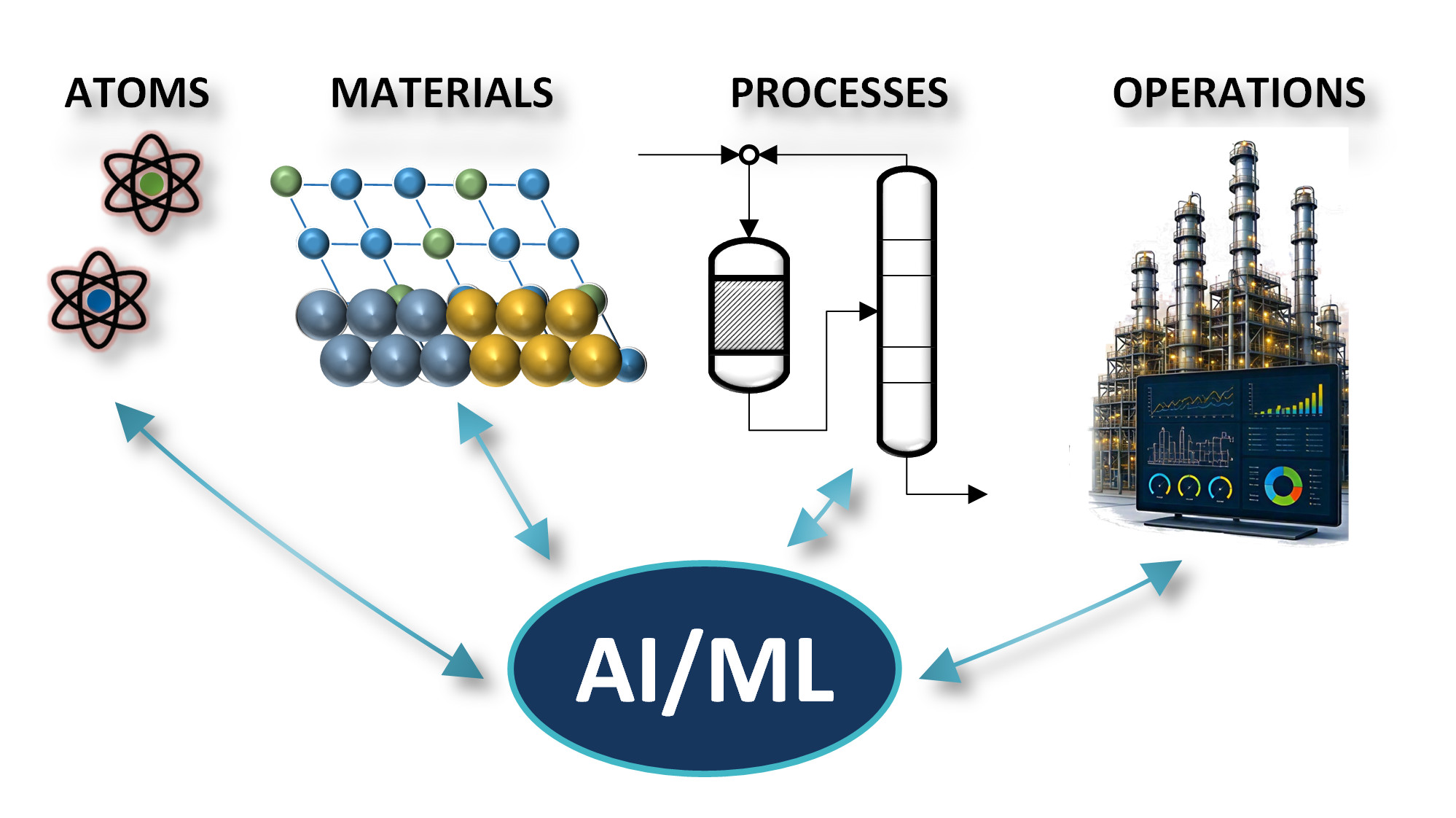}
\caption{For Table of Contents Only}
\end{figure*}

\end{document}